%% file: eccv2020submission.tex
\documentclass[runningheads]{llncs}
\usepackage{graphicx}
\usepackage{tikz}
\usepackage{comment}
\usepackage{amsmath,amssymb} 
\usepackage{color}
\usepackage[ruled,vlined]{algorithm2e}
\usepackage{url}

\usepackage{subcaption}

\makeatletter
\@namedef{ver@everyshi.sty}{}
\makeatother

\usepackage[colorinlistoftodos]{todonotes}

\newcommand{\name}{ConfigNet}
\newcommand{\synthdatasetname}{SynthFace}

\begin{document}
\pagestyle{headings}
\mainmatter
\def\ECCVSubNumber{1332}  

\title{CONFIG: Controllable Neural Face Image Generation} 

\titlerunning{CONFIG: Controllable Neural Face Image Generation}
%
\author{Marek Kowalski \and
Stephan J. Garbin \and
Virginia Estellers \and
Tadas Baltru\v{s}aitis \and
Matthew Johnson \and
Jamie Shotton}
\authorrunning{M. Kowalski et al.}
%
\institute{Microsoft}

\maketitle

\begin{abstract}
Our ability to sample realistic natural images, particularly faces, has
advanced by leaps and bounds in recent years, yet our ability to exert
fine-tuned control over the generative process has lagged behind. If
this new technology is to find practical uses, we need to achieve a
level of control over generative networks which,
without sacrificing realism, is on par with that seen in computer
graphics and character animation. To this end we propose \name{}, a
neural face model that allows for controlling individual aspects of
output images in semantically meaningful ways and that is a significant
step on the path towards finely-controllable neural rendering. \name{} is
trained on real face images as well as synthetic face renders.
Our novel method uses synthetic data to factorize the latent space
into elements that correspond to the inputs of a traditional rendering
pipeline, separating aspects such as head pose, facial expression,
hair style, illumination, and many others which are very hard to annotate
in real data. The real images, which are presented to the network without
labels, extend the variety of the generated images and encourage realism.
Finally, we propose an evaluation criterion using an attribute detection
network combined with a user study and demonstrate state-of-the-art
individual control over attributes in the output images.
\keywords{neural rendering; face image manipulation; GAN;}
\end{abstract}

\begin{figure}
  \centering
  \includegraphics[width=\textwidth]{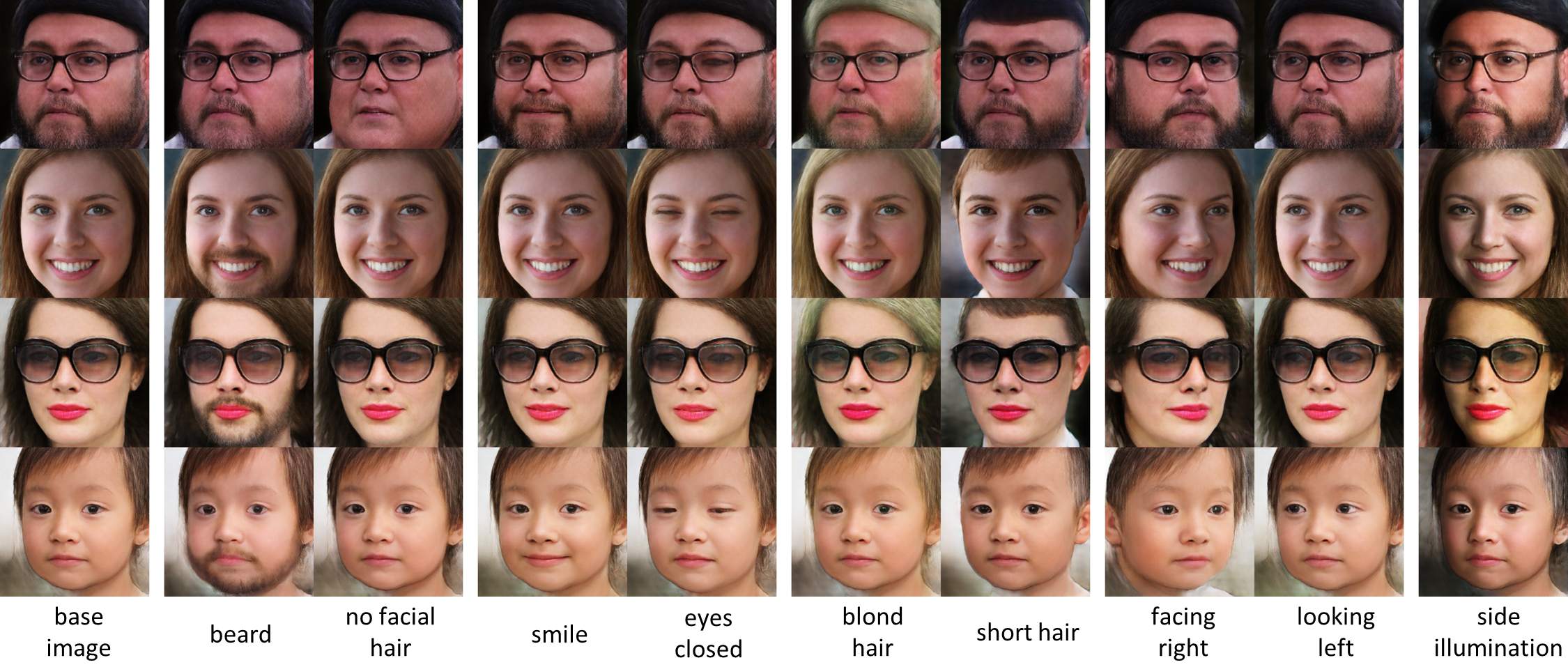}
  \caption{\name{} learns a factorized latent space, where each part corresponds to a different facial attribute. The first column shows images produced by \name{} for certain points in the latent space. The remaining columns show changes to various parts of the latent space vectors, where we can generate attribute combinations outside the distribution of the training set like children or women with facial hair.}
  \label{fig:intro_top}
  \end{figure}

\begin{figure}
 \centering
  \includegraphics[width=\textwidth]{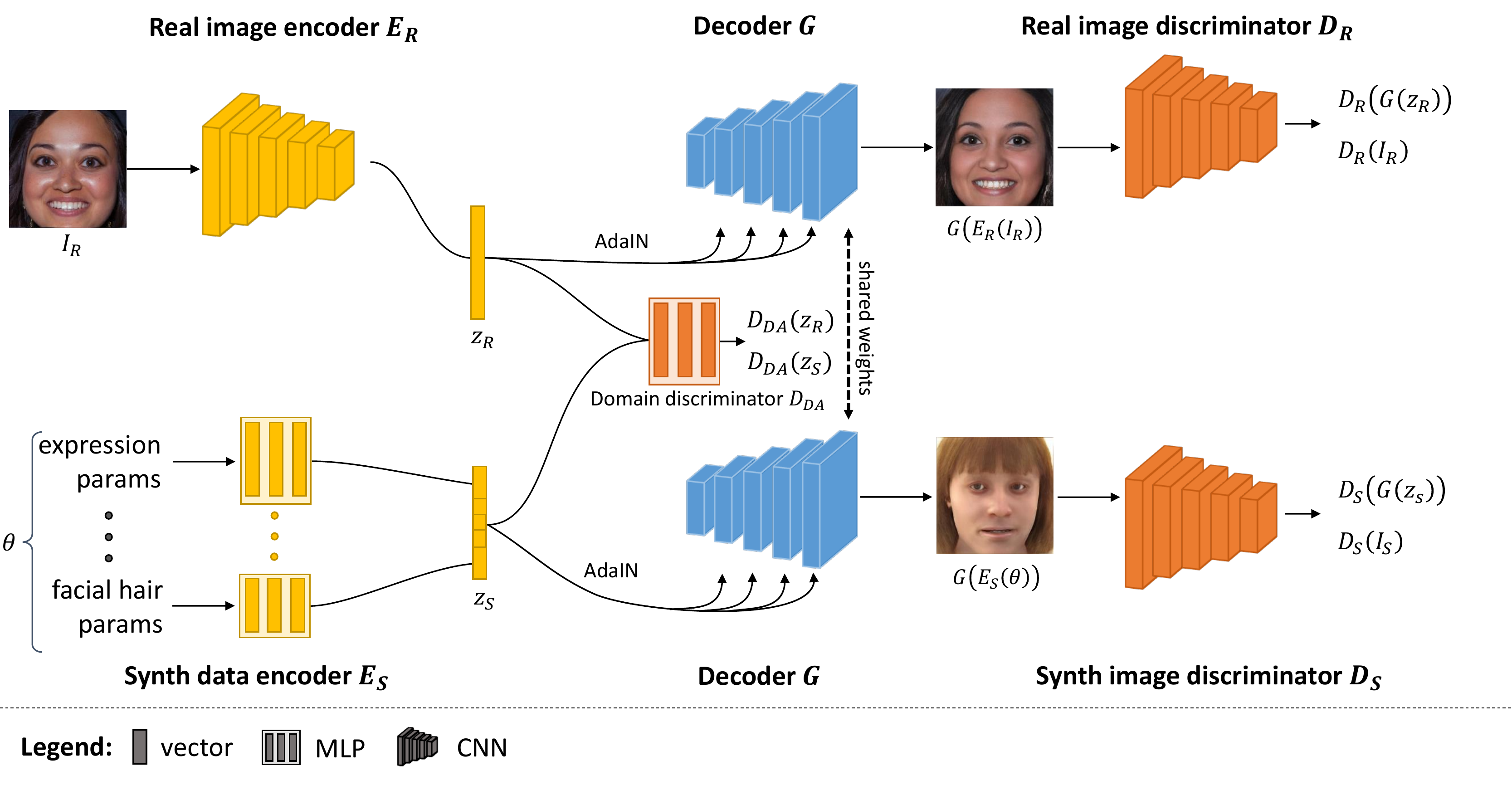}
  \caption{\name{} has two encoders $E_R$ and $E_S$ that encode real face images $I_R$ and the parameters $\theta$ of synthetic face images $I_S$. The encoders output latent space vectors $z_R$, $z_S$. The shared decoder, $G$, generates both real and synthetic images. A domain discriminator $D_{DA}$ ensures the latent distributions generated by $E_R$ and $E_S$ are similar.}
  \label{fig:outline}
\end{figure}

\section{Introduction}
Recent advances in generative adversarial networks (GANs) \cite{StyleGAN,StyleGAN2,BigGAN} have enabled the production of realistic high resolution images of smooth organic objects such as faces.
Generating photorealistic human bodies, and faces in particular, with traditional rendering pipelines is notoriously difficult \cite{mori1970uncanny}, requiring hand-crafted 3D assets. However, once these assets have been generated we can render the face from any direction and in any pose. In contrast, GANs can be used to easily generate realistic head and face images without the need to author expensive 3D assets, by training on curated datasets of 2D images of real human faces. However, it is difficult to enable meaningful control over this generation without detailed hand labelling of the dataset. Even when conditional models are trained with detailed labels, they struggle to generalize to out-of-distribution combinations of control parameters such as children with extensive facial hair or young people with gray hair. In order for GAN based rendering techniques to replace traditional rendering pipelines they must enable a greater level of control.

In this paper we present \name{}, one of the first methods to enable control of GAN outputs using the same methods as traditional graphics pipelines. The key idea behind \name{} is to train the generative model on both real and synthetically generated face images.
Since the synthetic images were generated with a traditional graphics pipeline, the renderer parameters for those images are readily available.
We use those known correspondences to train a generative model that uses the same input parametrization as the graphics pipeline used to generate the synthetic data. This allows for independent control of various face aspects including: head pose, hair style, facial hair style, expression and illumination.
By simultaneously training the model on unlabelled face images, it learns to generate photorealistic looking faces, while enabling full control over these outputs. Figure \ref{fig:intro_top} shows example results produced by \name{}.

\name{} can be used to both sample novel images and to embed existing ones, which can then be manipulated.
The ability to embed face images sets \name{} apart from traditional graphics pipelines, which would require person-specific 3D assets to achieve similar results. The use of a parametrization derived from a traditional graphics pipeline makes \name{} easy to use for people familiar with digital character animation.
For example, facial expressions are controlled with blendshapes with values in (0, 1), head pose is controlled with Euler angles and illumination can be set using an environment map.

Our main contributions are:
%
\begin{enumerate}
  \item \name{}, a novel method for placing real and synthetic data into a single factorized and disentangled latent space that is parametrized based on a computer graphics pipeline.
  \item A method for using \name{} to modify existing face images in a fine-grained way that allows for changing parts of the latent space factors meaningfully.
  \item Experiments showing our method generating realistic face images with attribute combinations that are not present in the real images of the training set. For example, a face of a child with extensive facial hair.
\end{enumerate}

\section{Related work}

\textbf{Image generation driven by 3D models}
One of the uses of synthetic data in image generation is the ``synthetic to real'' scenario, where the goal is to generate realistic images that belong to a target domain based on synthetic images, effectively increasing their realism.
The methods that tackle this problem \cite{SimGAN,FaceGAN,CycleGAN} usually use a neural network with an adversarial and semantic loss to push a synthetic image closer to the real domain.
While those type of methods can generate realistic images that are controllable through synthetic data, the editing of existing images is difficult as it would require fitting the underlying 3D model to an existing image.

3D model parameters can also be a supervision signal for generative models as shown in \cite{shen2018faceid,kulkarni2015deep} and most recently StyleRig \cite{StyleRig}.
This group of methods shares the challenges of the synthetic to real scenario as they require fitting the 3D model to existing images.

PuppetGAN \cite{PuppetGAN}, is a method designed to edit existing images using synthetic data of the same class of objects.
It uses two encoder-decoder pairs, one for real and one for synthetic images, which have a common latent space, part of which is designated for an attribute of interest that can be controlled. An image can be edited by encoding it with the real-data encoder, then swapping the attribute of interest part of the latent space with one encoded from a synthetic image and finally decoding with the real-data decoder. Due to the use of separate decoders for real and synthetic data PuppetGAN struggles to decode images where the attribute of interest is outside of the range seen in real data. The method performs well for a single attribute. In contrast, \name{} demonstrates disentanglement of multiple face attributes as well as generation of attribute combinations that do not exist in the real training data.

\textbf{Disentangled representation learning}
Supervised disentanglement methods try to learn a factorised representation, parts of which correspond to some semantically meaningful aspects of the generated images, based on labelled data in the target domain (as opposed to the synthetic data domain). The major limitation of these methods \cite{Mathieu16,NeuralTalkingHead,MakeaFace,choi2018stargan} is that they are only able to disentangle  factors of variations that are labelled in the training set. For human faces, labels are easily obtainable for some attributes, such as identity, but the task becomes more difficult with attributes like illumination and almost impossible with attributes like hair style. This labelling problem also becomes more difficult as the required fidelity of the labels increases (e.g. smile intensity).

Unsupervised disentanglement methods share the above goal but do not require labelled data. Most methods in this family, such as $\beta$-VAE \cite{BetaVAE}, InfoGAN \cite{InfoGAN}, ID-GAN \cite{ID-GAN}, place constraints on the latent space that lead to disentanglement. The fundamental problem with those approaches is that there is little control over what factors get disentangled and which part of the latent space corresponds to a given factor of variation. HoloGAN \cite{HoloGAN} separates the 3D rotation of the object in the image from variation in its shape and appearance. \name{} borrows the generator architecture of HoloGAN, while disentangling many additional factors of variation and allowing existing images to be edited.

Detach and Adapt \cite{liu2018detach} is trained in a semi-supervised way on images from two related domains, only one of which has labels. The resulting model allows generating images in both domains with some control over the labelled attribute.

\textbf{Face video re-enactment}
Face video re-enactment methods aim to produce a video of a certain person's articulated face that is driven by a second video of the same or a different person. The methods in \cite{Face2Face,NeuralTalkingHead,X2Face} achieve some of the most impressive face manipulation results seen to date.
Face2Face \cite{Face2Face} fits a 3D face model and illumination parameters to a video of a person and then re-renders the sequence with modified expression parameters that are obtained from a different sequence. This approach potentially allows for modifying any aspects of the rendered face that can be modelled, rendered and fitted to the input video. In practice, due to limitations of existing 3D face models and fitting methods, this approach cannot modify complex face attributes like hair style or attributes that require modelling of the whole head, like head pose.

Zakharov et al. \cite{NeuralTalkingHead} propose a video re-enactment method where the images are generated by a neural network driven by face landmarks from a different video sequence. The method produces impressive results given only a small number of target frames. X2Face \cite{X2Face} uses one neural network to resample the source image into a standard reference frame and a second network that resamples this standardized image into a different head pose or facial expression, which can be driven by images or audio signal.
While these two methods produce convincing results, the controllability is limited to head pose and expression. 

\section{Method}

The key concept behind the proposed method is to factorize the latent space into parts that correspond to separate and clearly-defined aspects of face images. These factors can be individually swapped or modified (Section \ref{sec:fine_grained_control}) to modify the corresponding aspect of the output image. The factorization needs labels that fully explain the image content, which would require laborious annotation for real data, but are easily obtained for synthetic data. We thus propose a generative model trained in a semi-supervised way, with labels that are known for synthetic data only. Figure \ref{fig:outline} outlines the proposed architecture.

\textbf{Overview}
Our approach is to treat the synthetic images $\mathcal{I}_S$ and real images $\mathcal{I}_R$ as two different subsets of a larger set of all possible face images.
Hence, the proposed method consists of a decoder $G$ and two encoders $E_R$ and $E_S$ that embed real and synthetic data into a common factorized latent space $z$ (Section \ref{sec:factorized_z}). We will refer to $z$ predicted by $E_R$ and $E_S$ as $z_R$ and $z_S$ respectively. The real data is supplied to its encoder as images $I_R\in\mathcal{I}_R$, while the synthetic data is supplied as vectors $\theta \in\mathbb{R}^m$ that fully describe the content of the corresponding image $I_S\in\mathcal{I}_S$. To increase the realism of the generated images we employ two discriminator networks $D_R$ and $D_S$ for real and synthetic data respectively.

We assume that the synthetic data is a reasonable approximation of the real data so that $\mathcal{I}_S \cap \mathcal{I}_R \neq \emptyset$.
Hence, it is desirable for $E_S(\Theta)$ and $E_R(\mathcal{I}_R)$, where $\Theta$ is the space of all $\theta$, to also be overlapping.
To do so, we introduce a domain discriminator network $D_{DA}$ and train it with a domain adversarial loss \cite{tzeng2017adversarial} on $z$, that forces $z_R$ and $z_S$ to be close.
In Section \ref{sec:ablation_study} we show that this loss is crucial for the method's ability to control the attributes of the output images.

To accurately reproduce and modify existing images we employ one-shot learning (Section \ref{sec:fine_tuning}) that improves reconstruction accuracy compared to embedding using $E_R$. To enable the sampling of novel images we train a latent GAN that generates samples of $z$ (Section \ref{sec:latent_gan}). Finally, we propose a method for modifying attributes of existing images in a fine-grained way that allows for changing parts of individual factors of $z$ meaningfully (Section \ref{sec:fine_grained_control}).

\subsection{Factorized latent space}
\label{sec:factorized_z}
Each synthetic data sample $\theta$ is factorised into $k$ parts $\theta_1$ to $\theta_k$, such that:
\begin{align}
  \theta \in\mathbb{R}^m = \mathbb{R}^{m_1} \times \mathbb{R}^{m_2} \times \ldots \times \mathbb{R}^{m_k}.
\end{align}
Each $\theta_i$ corresponds to semantically meaningful input of the graphics pipeline used to generate $\mathcal{I}_S$.
Examples of such inputs are: facial expression, facial hair parameters, head shape, environment map, etc.
The synthetic data encoder $E_S$ maps each $\theta_i$ to $z_i$, a part of $z$, which thus factorizes $z$ into $k$ parts.

The factorized latent space is a key feature of \name{} that allows for easy modification of various aspects of the generated images. For example, one might encode a real image into $z$ using $E_R$ and then change the illumination by swapping out the part of $z$ that corresponds to illumination. Note that the part of $z$ that is swapped in might come from $\theta_i$ (encoded by $E_S$), which is semantically meaningful, or it may come from a different real face image encoded by $E_R$.

\subsection{Loss functions}
To ensure that the output image $G(z)$ is close to the ground truth image $I_{GT}$, we use the perceptual loss $\mathcal{L}_{perc}$ \cite{perceptual_loss}, which is the MSE between the activations of a pre-trained neural network computed on $G(z)$ and $I_{GT}$. We use VGG-19 \cite{VGG} trained on ImageNet \cite{russakovsky2015imagenet} as the pre-trained network. We experimented with using VGGFace \cite{VGGFace} as base for the perceptual loss, but didn't see improvement.

While the perceptual loss retains the overall content of the image well, it struggles to preserve some small scale features. Because of that, we use an additional loss with the goal of preserving the eye gaze direction:
\begin{align}
\mathcal{L}_{eye} = w_{M}\sum M \circ (I_{GT} - G(z_s))~~\text{with}~~ w_{M} = (1 + \vert M\vert_{1})^{-1},
\end{align}
where $M$ is a pixel-wise binary mask that denotes the iris, only available for $\mathcal{I}_S$. Thanks to the accurate ground truth segmentation that comes with the synthetic data, similar losses could be added for any part of the face if necessary.

We train the adversarial blocks with the non-saturating GAN loss \cite{goodfellow2014generative}:
\begin{align}
  \mathcal{L}_{GAN_D}(D, x, y) &= \log D(x) + \log (1 - D(y)),\\
  \mathcal{L}_{GAN_G}(D, y)  &= \log (D(y)),
\end{align}
where $\mathcal{L}_{GAN_D}$ is used for the discriminator and $\mathcal{L}_{GAN_G}$ is used for the generator, $D$ is the discriminator, $x$ is a real sample and $y$ is a network output.

\subsection{Two-stage Training procedure}
\label{sec:training_procedure}
\textbf{First stage}: we train all the sub-networks except $E_R$, sampling $z_R\sim \mathcal{N}(0, \boldsymbol{I})$ as there is no encoder for real data at this stage. At this stage $E_S$ and $G$ are trained with the following loss:
\begin{align}
  \mathcal{L}_1 & = \mathcal{L}_{GAN_G}(D_R, G(z_R)) + \mathcal{L}_{GAN_G}(D_{DA}, z_S) \nonumber \\
  & + \mathcal{L}_{GAN_G}(D_S, G(z_S))  + \lambda_{eye} \mathcal{L}_{eye} + \lambda_{perc} \mathcal{L}_{perc}\left( G(z_S), I_S \right),
\end{align}
where $z_S = E_S(\theta)$ and $\lambda$ are the loss weights. The domain discriminator $D_{DA}$ acts on $E_S$ to bring the distribution of its outputs closer to $\mathcal{N}(0, \boldsymbol{I})$ and so $E_S$ effectively maps the distribution of each $\theta_i$ to $\mathcal{N}(0, \boldsymbol{I})$.

\textbf{Second stage}: we add the real data encoder $E_R$ so that $z_R = E_R(I_R)$. The loss used for training $E_S$ and $G$ is then:
\begin{align}
  \mathcal{L}_2 = \mathcal{L}_1 + \lambda_{perc} \mathcal{L}_{perc}\left( G(z_R), I_R \right) + \log (1 - D_{DA}(z_R)),
\end{align}
where the goal of $\log (1 - D_{DA}(z_R))$ is to bring the output distribution of $E_R$ closer to that of $E_S$. In the second stage we increase the weight of $\lambda_{perc}$, in the first stage it is set to a lower value as otherwise the total loss for synthetic data would overpower that for real data. Our experiments show that this two-stage training improves controllability and image quality.

\subsection{One-shot learning by fine-tuning}
\label{sec:fine_tuning}
Our architecture allows for embedding face images into $z$ using the real data encoder $E_R$, individual factors $z_i$ can then be modified to modify the corresponding output image as explained in Sections \ref{sec:factorized_z} and \ref{sec:fine_grained_control}. We have found that while $G(E_R(I_R))$ is usually similar to $I_R$ as a whole image, there is often an identity gap between the face in $I_R$ and in the generated image. A similar finding was made in \cite{NeuralTalkingHead}, where the authors proposed to decrease the identity gap by fine-tuning the generator on the images of a given person.

Similarly, we fine-tune our generator on $I_R$ by minimizing the following loss:
\begin{align}
  \mathcal{L}_{ft} & = \mathcal{L}_{GAN_G}(D_R, G(\hat{z_R})) + \log (1 - D_{DA}(\hat{z_R})) \nonumber \\
  & + \lambda_{perc} [\mathcal{L}_{perc}\left( G(\hat{z_R}), I_R \right) + \mathcal{L}_{face}\left( G(\hat{z_R}), I_R \right)],
\end{align}
where $\mathcal{L}_{face}$ is a perceptual loss with VGGFace \cite{VGGFace} as the pre-trained network. We optimize over the weights of $G$ as well as $\hat{z_R}$ which is initialized with $E_R(I_R)$. The addition of a $\mathcal{L}_{face}$ improves the perceptual quality of the generated face images. We believe that this improvement is visible here, but not in the main training phase, as fine-tuning lacks the regularization provided by training on a large number of images and can easily ``fool'' the single perceptual loss.

\subsection{Sampling of $z$}
\label{sec:latent_gan}
While the proposed method allows for embedding existing face images into the latent space, sometimes it might be desirable to sample the latent space itself. Samples of the latent space can be used to generate novel images or to sample individual factor $z_i$. The sampled $z_i$ can then be used to generate additional variations of an existing image that was embedded in $z$.

To do this, we use a latent GAN \cite{LatentGAN}. The latent GAN is trained to map between its input $w \sim \mathcal{N}(0, \boldsymbol{I})$ and the latent space $z$. This simple approach allows for sampling the latent space without the constraints on $z$ imposed by VAEs that lead to reduced quality. The latent GAN is trained with the GAN losses described above, both the discriminator and generator $G_{lat}$ are 3-layer MLPs. Figure 18
in suppl. shows an outline of the method when used with $G_{lat}$.

\subsection{Fine-grained control}
\label{sec:fine_grained_control}
Given an existing face image embedded into $z$, we can easily swap any part, $z_i$, of its embedding with one that is obtained from $E_S$ or $E_R$. However, sometimes we might want a finer level of control and only modify a single aspect of $z_i$ while leaving the rest the same. If $z_i$ is a face expression, its single aspect might be the intensity of smile, if $z_i$ is illumination, the brightness might be one aspect. These aspects are controlled by individual elements of the corresponding $\theta_i$ vector. However $\theta_i$ is unknown if $z$ was generated by $E_R$ or $G_{lat}$.

For this reason, we use an approximation $\tilde{\theta}_i$ obtained by solving the minimization problem $\min_{\tilde{\theta}_i} \vert z_i - E_{S_i}(\tilde{\theta}_i)\vert^2$ with gradient descent, where $E_{S_i}$ is the part of $E_S$ that corresponds to $\theta_i$.
We incorporate constraints on $\theta_i$ into the optimization algorithm. For example, our expression parameters lie in the convex set $[0, 1]$ and we use projected gradient descent to incorporate the constraint into the minimization algorithm. Given $\tilde{\theta}_i$, e.g. a face expression vector, we can modify the part of the vector responsible for an individual expression and use $E_S$ to obtain a new latent code $z_i$ that generates images where only this individual expression is modified. We use this approach to manipulate individual expressions in Figures \ref{fig:intro_top}, \ref{fig:puppetgan_comp_mouth_opening} and combinations in Figure 12
(supplementary). The method described above is also outlined in Algorithm 1
in supplementary.

\subsection{Implementation}
\label{sec:implementation}
The architecture of the decoder $G$ is based on the generator used in HoloGAN \cite{HoloGAN}, explained in supplementary. We choose this particular architecture as it decouples object rotation from $z$ and it allows for specifying the rotation with any parametrization. This lets us obtain the poses of the heads in $\mathcal{I}_S$ in \name{} parametrization and supply head pose directly, without an encoder.

The remaining $k-1$ parts of $\theta$ are encoded with separate multi layer perceptrons (MLPs) $E_{S_i}$, each of which consists of 2 layers with number of hidden units equal to the dimensionality of the corresponding $\theta_i$. The real image encoder $E_R$ is a ResNet-50 \cite{ResNet} pre-trained on ImageNet \cite{Imagenet}. The domain discriminator $D_{DA}$ is a 4-layer MLP. The two image discriminators $D_R$ and $D_S$ share the same basic convolutional architecture. The supplementary material contains all network details, source code is available at \url{http://aka.ms/confignet}.

\section{Experiments}
\label{sec:experiments}
\textbf{Datasets}
We use the FFHQ \cite{StyleGAN} (60k images, 1Mpix each), and \synthdatasetname{} (30k images, 1Mpix each) datasets as a source of real and synthetic training images. We align the face images from all datasets to a standard reference frame using landmarks from OpenFace \cite{OpenFace,OpenFaceLandmarks1,OpenFaceLandmarks2} and reduce the resolution to 256x256 pixels.

\begin{figure}[t]
  \centering
  \includegraphics[width=\textwidth]{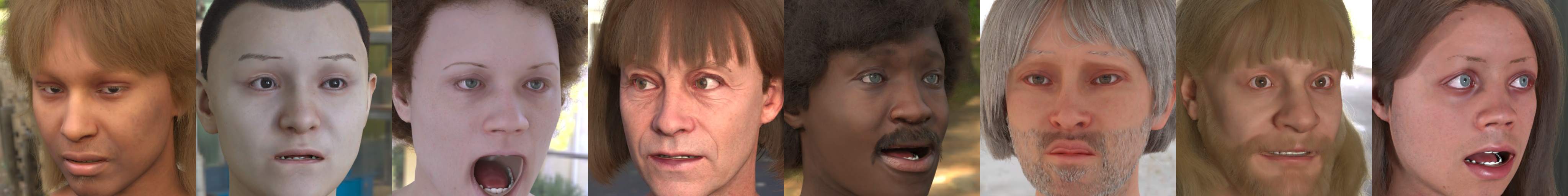}
  \caption{Images from \synthdatasetname{} dataset, note the domain gap to real images.}
  \label{fig:synth_data}
\end{figure}

Our experiments use the 10k images in the validation set of FFHQ to evaluate \name{}. The \synthdatasetname{} dataset was generated using the method of \cite{SynthFace} and setting rotation limits for yaw and pitch to $\pm 30^{\circ}$ and $\pm 10^{\circ}$ to cover the typical range of poses in face images. For \synthdatasetname{}, $\theta$ has $m=304$ dimensions, while $z$ has $n=145$ dimensions, and is divided into $k=12$ factors. Table 6
in the supplementary provides the dimensionality of each factor in $\theta$ and $z$, Figure \ref{fig:synth_data} shows sample \synthdatasetname{} images.

\subsection{Evaluation of \name{}}
\label{sec:evaluation}
Our experiments evaluate \name{} key features: photorealism and control.

\begin{figure}[t]
  \centering
  \includegraphics[width=0.4\textwidth]{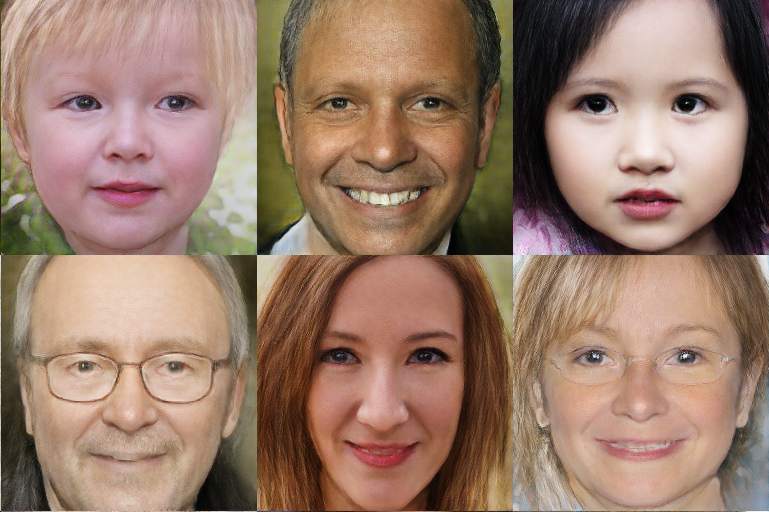}
  \quad
  \includegraphics[width=0.4\textwidth]{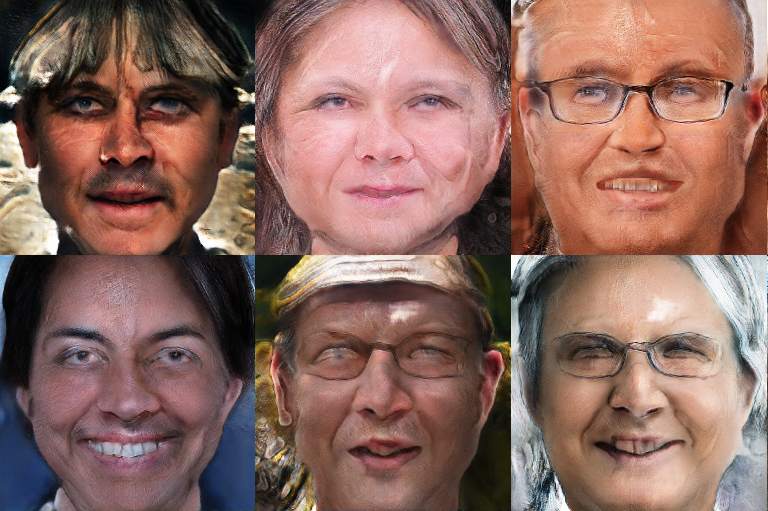}
  \caption{Left: $G(z)$ trained using the two-stage method, where $z$ is sampled from latent GAN. Right: $G(z)$ trained using the first stage only, where $z$ is sampled from prior. Note the large improvement in quality when second stage is added.}
  \label{fig:latent_gan}
\end{figure}

\textbf{Photorealism} Figure \ref{fig:latent_gan} shows samples generated by the latent GAN (where $E_R, G$ were trained using the two stage-procedure of Section \ref{sec:training_procedure}) and a standard GAN model trained only with the first-stage procedure. We observe a large improvement in photorealism when the second stage of training is added. We believe that the low-quality images produced by the standard GAN are caused by the constraint $z \sim \mathcal{N}(0, \boldsymbol{I})$, which is relaxed in our second-stage training thus allowing real and synthetic data to co-exist in the same space.

We quantitatively measure the photorealism and coverage of the generated images using the Frechet Inception Distance (FID) \cite{FID} in Table \ref{tab:fid_kid}. The latent GAN achieves scores that are close to those produced by sampling $z$ through $E_R$, which is the upper limit of its performance. Training only the first stage and sampling $z \sim \mathcal{N}(0, \boldsymbol{I})$ results in poorer metrics. As expected, the raw synthetic images give the worst result. To further evaluate how much of the photorealism of the generated data is lost due to training on both real and synthetic data, we train \name{} without synthetic data and the losses that require its presence. We find that the resulting FID is very close to those produced by our standard training. This suggests that the photorealism of the results might be limited by our network architecture rather then by the use of synthetic data. We speculate that using a more powerful $G$ and $D_R$, for example the ones used in StyleGAN, may lead to improved results.

\begin{table}[t!]
  \caption{FID score for FFHQ, \synthdatasetname{}, and images obtained with our decoder $G$ and latent codes from the real-image encoder $E_R$ and latent GAN $G_{lat}$. }
  \centering
  \begin{tabular}{ l l l }
    Method & FID$\downarrow$ \\
    \hline
    $G(E_R(I_R))$ & 33.41 \\
    synthetic data $\mathcal{I_S}$ & 52.19 \\
    $G(E_R(I_R))$ trained without $\mathcal{I_S}$& 33.49 \\
    $G(G_{lat}(w)),~w \sim \mathcal{N}(0, \mathcal(I))$ & 39.76 \\
    $G(z),~z \sim \mathcal{N}(0, \mathcal(I))$ no 2nd stage & 43.05 \\
  \end{tabular}
  \label{tab:fid_kid}
\end{table}

\textbf{Controllability}
We evaluate \name{}'s controllability analysing how changing a specific attribute (e.g., hair colour) changes the output image: with perfect control, the output image should only change with respect to that attribute.

Figure \ref{fig:intro_top} and \ref{fig:fine_tuning} show controllability qualitatively. Figure \ref{fig:intro_top} shows that the generator is able to modify individual attributes of faces embedded in its latent space, while Figure \ref{fig:fine_tuning} shows that each attribute can take many different values while only influencing certain aspects of the produced image. The second column of Figure \ref{fig:intro_top} shows that we are able to set facial hair to faces of children and women, demonstrating that the generator is not constrained by the distribution of the real training data. Fine-grained control over individual expressions is shown in Figure \ref{fig:puppetgan_comp_mouth_opening} as well as Figure 12
in the supplementary. The supplementary also includes additional results of face attribute manipulation and interpolation, including a video.

To evaluate if \name{} offers this ideal level of control quantitatively, we propose the following experiment: We take a random image $I_R$ from the FFHQ validation set, encode it into latent space $z = E_R(I_R)$ and then swap the latent factor $z_i$ that corresponds to a given attribute $v$ (for example hair colour) with a latent factor obtained with $E_S$. For each attribute $v$ we output two images: $I_+$ where the attribute is set to a certain value $v_+$ (e.g. blond hair) and another $I_-$ with the attribute takes a semantically opposite value $v_-$ (e.g., black hair)\footnote{We choose the values of $\Theta_i$ for $v_+$ and $v_-$ by manual inspection, details in suppl.}. This gives us image pairs $(I_+, I_-)$ that should be identical except for the chosen attribute $v$, where they should differ. We measure how and where these images differ with an attribute predictor and a user study.

\begin{figure}[t]
\begin{subfigure}[t]{\textwidth}
\centering
\includegraphics[width=\textwidth]{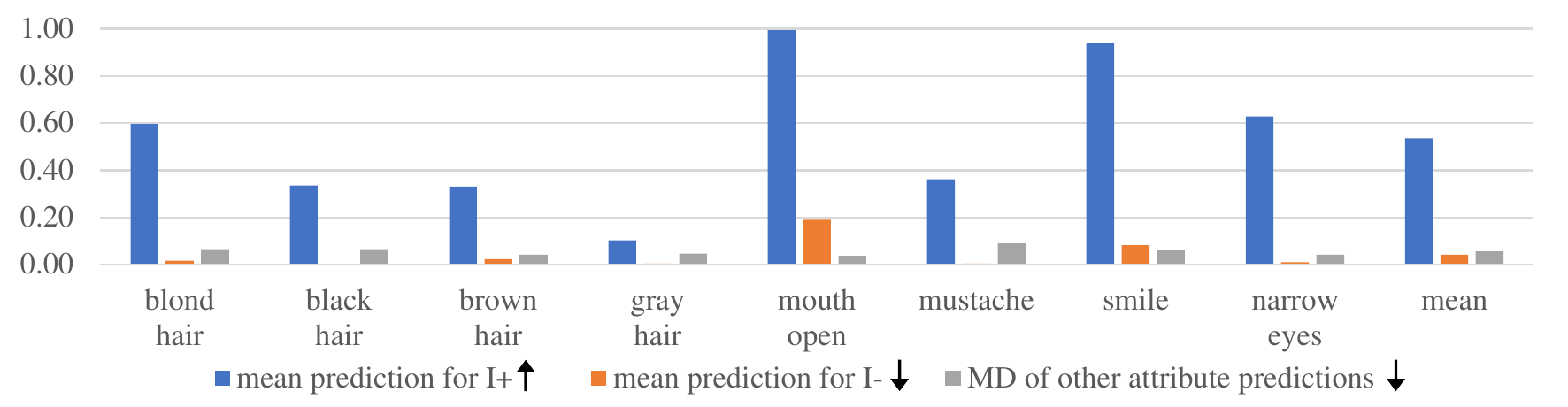}
\caption{Evaluation of controllability and disentanglement with an attribute predictor.}
\label{fig:controllability_predictor}
\end{subfigure}
\begin{subfigure}[t]{\textwidth}
\centering
  \includegraphics[width=\textwidth]{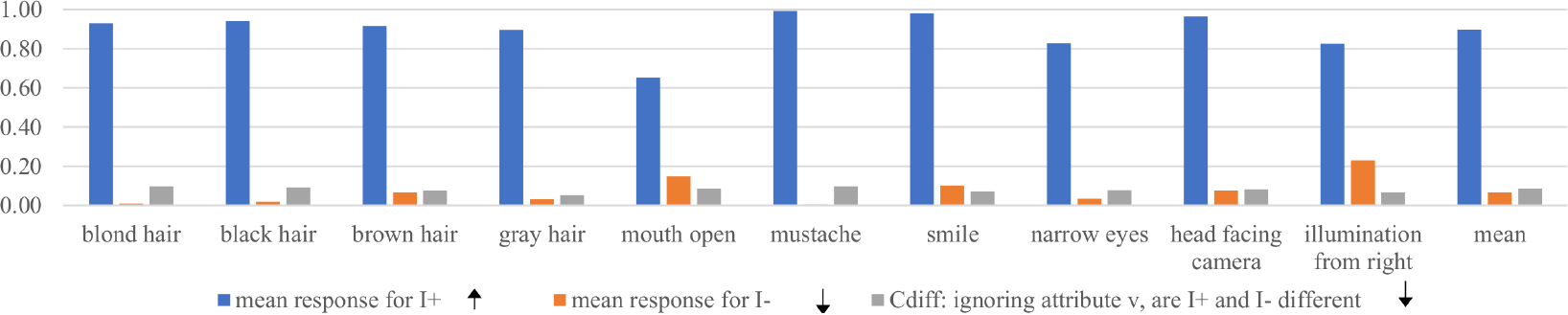}
\caption{Evaluation of controllability and disentanglement with a user study.}
\label{fig:controllability_user_study}
\end{subfigure}
  \caption{Evaluation of control and disentanglement of \name. Blue and orange bars show the predicted values of given attribute for images with that attribute ($I_+$, higher  better) and images with an opposite attribute ($I_-$, lower better). The gray bars measure differences of other attributes (MD and $C_{diff}$, lower better).}
\end{figure}

We train an attribute predictor $C_{pred}$ on CelebA \cite{CelebA} to predict 38 face attributes and use it with 1000 FFHQ validation images to estimate 1) if $v_+$ is present in each set of images pairs $(I_+, I_-)$ and 2) if the other face attributes change. Ideally, $C_{pred}(I_+) = 1, ~C_{pred}(I_-)=0$ and the Mean Absolute Difference (MD) for other face attributes should be 0. Figure \ref{fig:controllability_predictor} shows how $C_{pred}(I_+) \ggg C_{pred}(I_-)$ while the MD of other attributes is close to 0. The best controllability is achieved for the mouth opening and smile attributes, with $C_{pred}(I_+)$ approaching the ideal value of 1, while the poorest results are achieved for the gray hair attribute. We believe those large differences are caused by bias in CelebA, where certain attributes are not distributed evenly across age (for example gray hair) or gender (for example moustache).

Our user study $C_{user}$ follows a similar evaluation protocol: 59 users evaluated the presence of $v_+$ in a total of 1771 images pairs $I_+$ and $I_-$ on a 5-level scale and gave a score $C_{diff}$ that measures whether, ignoring $v$, the images depict the same person. Figure \ref{fig:controllability_user_study} shows the results of the controllability and disentanglement metrics for the user study: users evaluate the controllability of the given attribute higher than the feature predictor $C_{pred}$, with $C_{user}(I_+)>C_{pred}(I_{+})$ and $C_{user}(I_+)-C_{user}(I_-)> C_{pred}(I_+)-C_{pred}(I_-)$ for all features except \textit{mouth open}, while the score $C_{diff}$ measuring whether $I_+$ and $I_-$ show the same person has low values indicating that features other than $v_{+}$ remain close to constant. This results support the result of the feature predictor and show a similar performance for different attributes because user judgements do not suffer from the bias of the attribute predictor trained on CelebA.


\subsection{Ablation study}
\label{sec:ablation_study}
We evaluate the importance of two stage training and the domain discriminator $D_{DA}$ by training the neural network without them. Table \ref{tab:controllability} shows how each of those procedures contributes to controllability of \name{}. Compared to the base method, $C_{pred}(I_+) - C_{pred}(I_-)$ decreases by 60\% when the domain adversarial loss is removed and by 42\% when the first stage training is removed. Quantitatively, the mean absolute difference of the non-altered attributes, MD, is slightly larger for the base method. While this might seem a degradation caused by two stage training and the domain discriminator, we attribute the lower MD to the reduced capability of the network to modify the output image.

\begin{figure}[t]
  \centering
  \includegraphics[width=\textwidth]{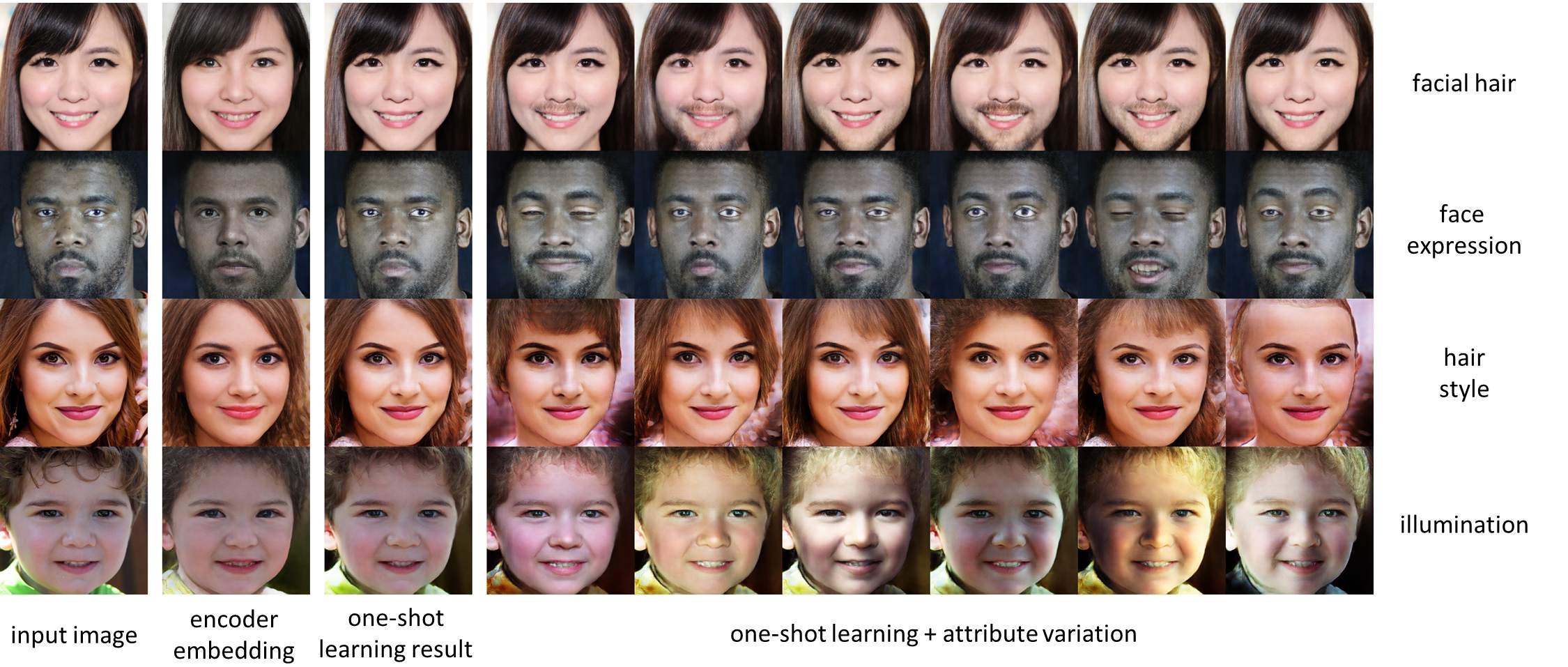}
  \caption{Effects of fine tuning and attribute variety. The first 3 columns show the input image, the results of the encoder embedding and fine tuning. the other columns show different facial attributes controllable modifying $E_{S_i}(\theta_i)$.}
  \label{fig:fine_tuning}
\end{figure}

One worry with fine-tuning\footnote{In all fine-tuning experiments we ran the fine-tuning procedure for 50 iterations.} on a single image is that it will change the decoder in a way that negatively affects controllability of the output image. Our experiments show that fine tuning leads to a 6\% reduction in $C_{pred}(I_+) - C_{pred}(I_-)$ and no increase of MD, which leads us to believe that the controllability of the fine-tuned generator is not significantly affected. Figure \ref{fig:fine_tuning} qualitatively shows the effects of fine-tuning compared to embedding using $E_R$.

An additional ablation study showing the influence of the eye gaze preserving loss $\mathcal{L}_{eye}$ is shown in Figure 16
in the supplementary.

\begin{table}[t!]
  \caption{Average controllability metrics for different variants of \name{}. $D_{DA}$ denotes the domain discriminator. Ideally, $C_{pred}(I_+)=1, ~C_{pred}(I_{-})=0$ and MD should be 0. The mean difference $C_{pred}(I_+) - C_{pred}(I_-)$ gives the dynamic range of a given attribute, the higher it is the more controllable the attribute.}
  \centering
  \begin{tabular}{ | l | c | c | c | c | }
    \hline
    Method & $C_{pred}(I_+)\uparrow$ & $C_{pred}(I_{-})\downarrow$ & MD$\downarrow$ & $C_{pred}(I_+) - C_{pred}(I_-)\uparrow$\\
    \hline
    base method & 0.54 & 0.04 & 0.06 & 0.50\\
    with fine-tuning & 0.52 & 0.05 & 0.05 & 0.47\\
    without $D_{DA}$ & 0.39 & 0.19 & 0.03 & 0.20\\
    without 1st stage & 0.43 & 0.14 & 0.04 & 0.29\\
    \hline
  \end{tabular}
  \label{tab:controllability}
\end{table}

\subsection{Comparison to state of the art}
\begin{figure}[t]
  \centering
  \includegraphics[width=0.45\textwidth]{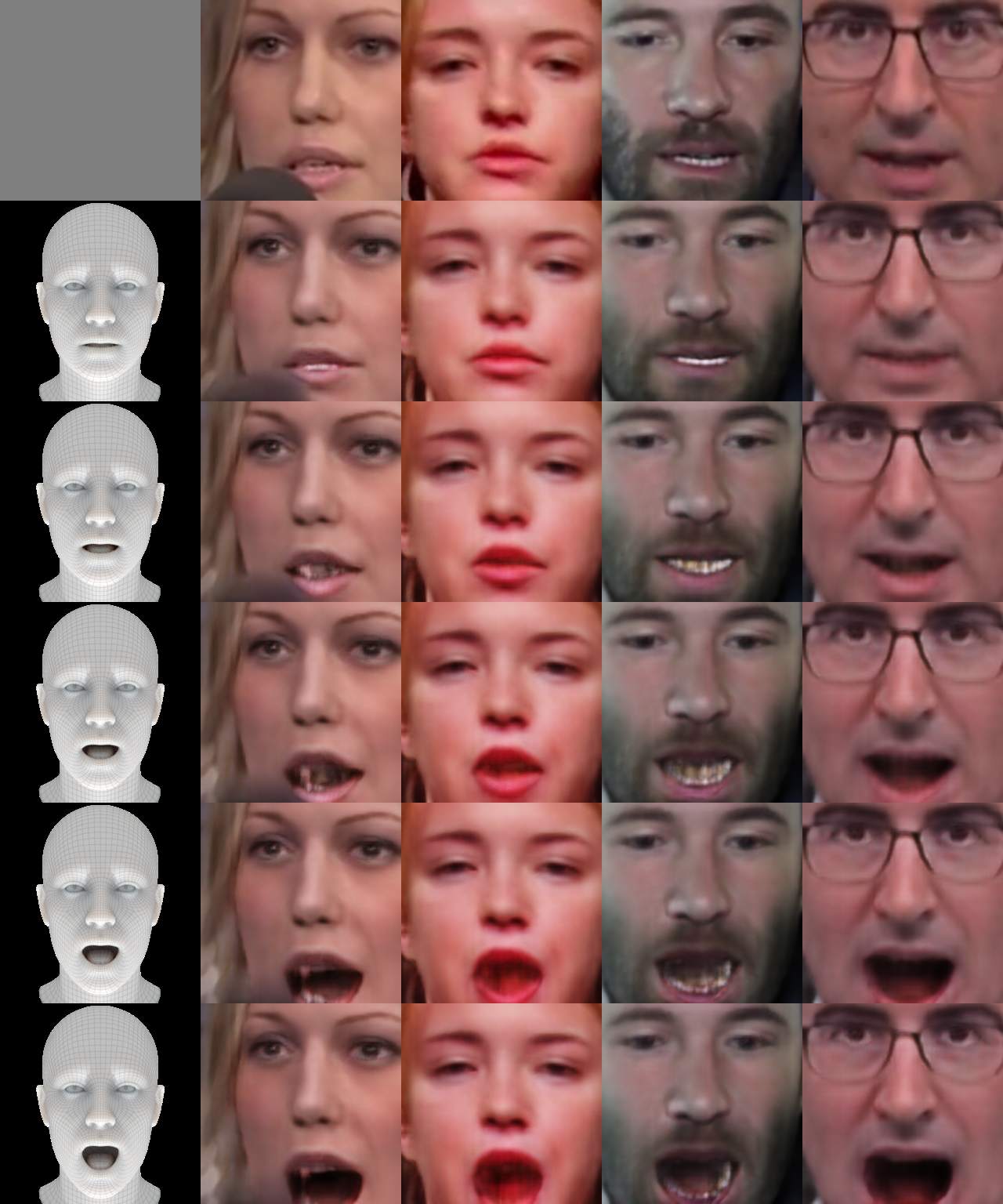}
  \includegraphics[width=0.45\textwidth]{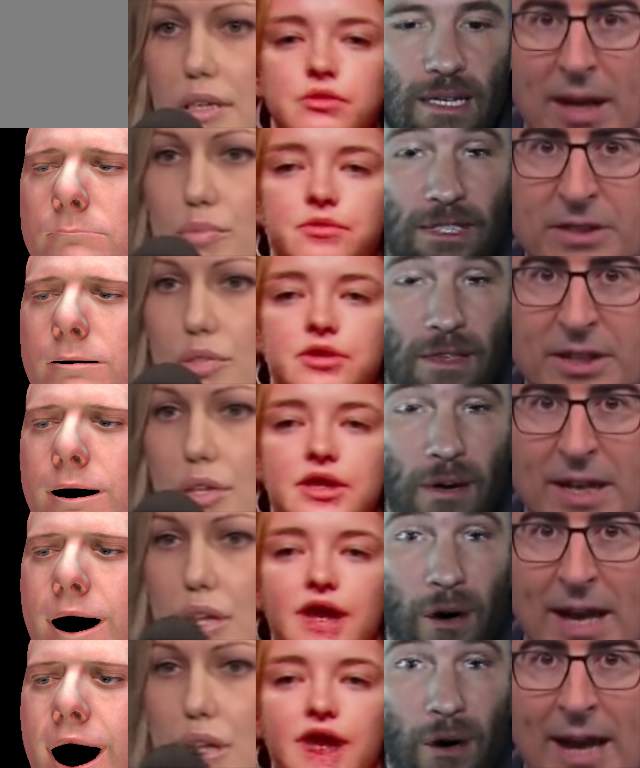}
  \caption{Comparison between \name{} (left) and PuppetGAN (right). Top row shows the input, left column the desired level of mouth opening for each row. To facilitate  comparison, \name{} results are cropped to match PuppetGAN.}
  \label{fig:puppetgan_comp_mouth_opening}
\end{figure}

In this section we compare to PuppetGAN \cite{PuppetGAN}, which is the most closely related method, additional comparisons to CycleGAN \cite{CycleGAN} and Face-ID GAN \cite{shen2018faceid} are in the supplementary. For comparison to PuppetGAN we use a figure from \cite{PuppetGAN} that shows control over the degree of mouth opening on frames from several videos from the 300-VW dataset \cite{300VW}. To generate the figure, the authors of PuppetGAN trained separate models on each of the videos and then demonstrated the ability to change the degree of mouth opening in the frames of the same video.

To generate similar results we use a model trained on FFHQ and fine-tune it on the input frame using the method described in Section \ref{sec:fine_tuning}. We then use the fine-grained control method (Section \ref{sec:fine_grained_control}) to change only the degree of mouth opening. The results of this comparison are shown in Figure \ref{fig:puppetgan_comp_mouth_opening}. At a certain level of mouth opening PuppetGAN saturates and is not able to open the mouth more widely, \name{} does so, while retaining a similar level of quality and disentanglement. Both methods fail to close the mouth fully for some of the input images. We believe that in case of \name{} this is an issue with the disentanglement of the synthetic training set itself, we give further details and describe a solution in supplementary materials. It is also worth noting that PuppetGAN uses hundreds of training images of a specific person, while \name{} requires only a single frame and it is able to modify many additional attributes.

\subsection{Failure modes}
\label{sec:failure_modes}

\begin{figure}[t]
  \centering
  \begin{subfigure}[t]{0.3\textwidth}
    \includegraphics[height=1.7cm]{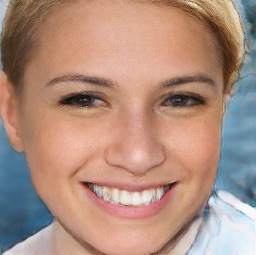}
    \includegraphics[height=1.7cm]{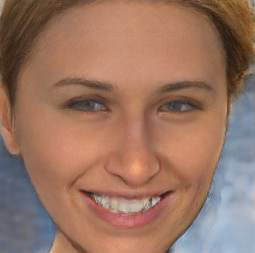}
  \end{subfigure}
  \begin{subfigure}[t]{0.3\textwidth}
    \centering
    \includegraphics[height=1.7cm]{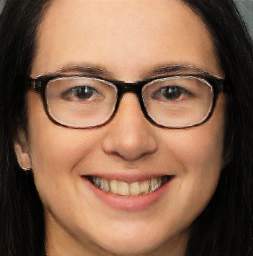}
    \includegraphics[height=1.7cm]{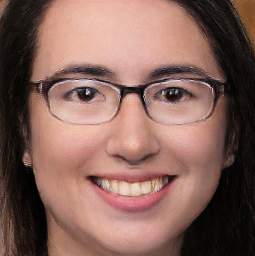}
  \end{subfigure}
  \begin{subfigure}[t]{0.15\textwidth}
    \includegraphics[height=1.7cm]{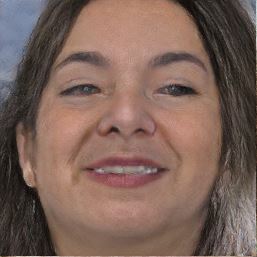}
  \end{subfigure}
  \caption{Failure modes. Left image pair: changing head shape to one obtained from $\theta$ moves the appearance of the image closer to synthetic data. Central image pair: change of $z_i$ corresponding to texture changes style of glasses. Right: frontal image generated from an image $I_R$ with pose outside the supported range.}
  \label{fig:failure_modes}
\end{figure}
One of the key issues we have identified is that the $z_i$ that corresponds to head shape is often separated for real and synthetic data. For example, changing the head shape of a real image embedded into $z$ using $E_S(\theta_i)$ results in the face appearing closer to the synthetic image space and some of its features being lost, see Figure \ref{fig:failure_modes}a for an example. This separation is placed in the head shape space very consistently, we believe this is because head shape affects the whole image in a significant way, so it's easy for the generator to ``hide'' the difference between real and synthetic images there.

Another issue is that \synthdatasetname{} does not model glasses, which leads to \name{} hiding the representation of glasses in unrelated face attributes, most commonly texture, head and eyebrow  shape, as shown in Figure \ref{fig:failure_modes}b. Lastly, we have found that when $I_R$ has a head pose that is out of the rotation range of \synthdatasetname{}, the encoder $E_R$ hides the rotation in other parts of $z$, as shown in Figure \ref{fig:failure_modes}c. We believe this is a result of constraining the rotation output of $E_R$ to the range seen in \synthdatasetname{} (details in supplementary). Generating a synthetic dataset with a wider rotation range would likely alleviate this issue.

\section{Conclusions}
We have presented \name{}, a novel face image synthesis method that allows for controlling the output images to an unprecedented degree. Crucially, we show the ability to generate realistic face images with attribute combinations that are outside the distribution of the real training set. This unique ability brings neural rendering closer to traditional rendering pipelines in terms of flexibility.

An open question is how to handle aspects of real face images not present in synthetic data. Adding additional variables in the latent space to model these aspects only for real data is an investigation that we leave to future work.

In the short term, we believe that \name{} could be used to enrich existing datasets with samples that are outside of their data distribution or be applied to character animation. In the long term, we hope that similar methods will replace traditional rendering pipelines and allow for controllable, realistic and person-specific face rendering.

\textbf{Acknowdledgments}
The authors would like to thank Nate Kushman for helpful discussions and suggestions.

\nocite{LFW,mescheder2018training,ulyanov2016instance,huang2017arbitrary,sandler2018mobilenetv2,Adam,ProgressiveGAN}

\bibliographystyle{splncs04}
\bibliography{egbib}

\clearpage
\input{supplementary_materials}
\end{document}

%% file: supplementary_materials.tex
\section{Supplementary}
\subsection{High-resolution image generation}
\begin{figure}
    \centering
    \includegraphics[width=0.95\textwidth]{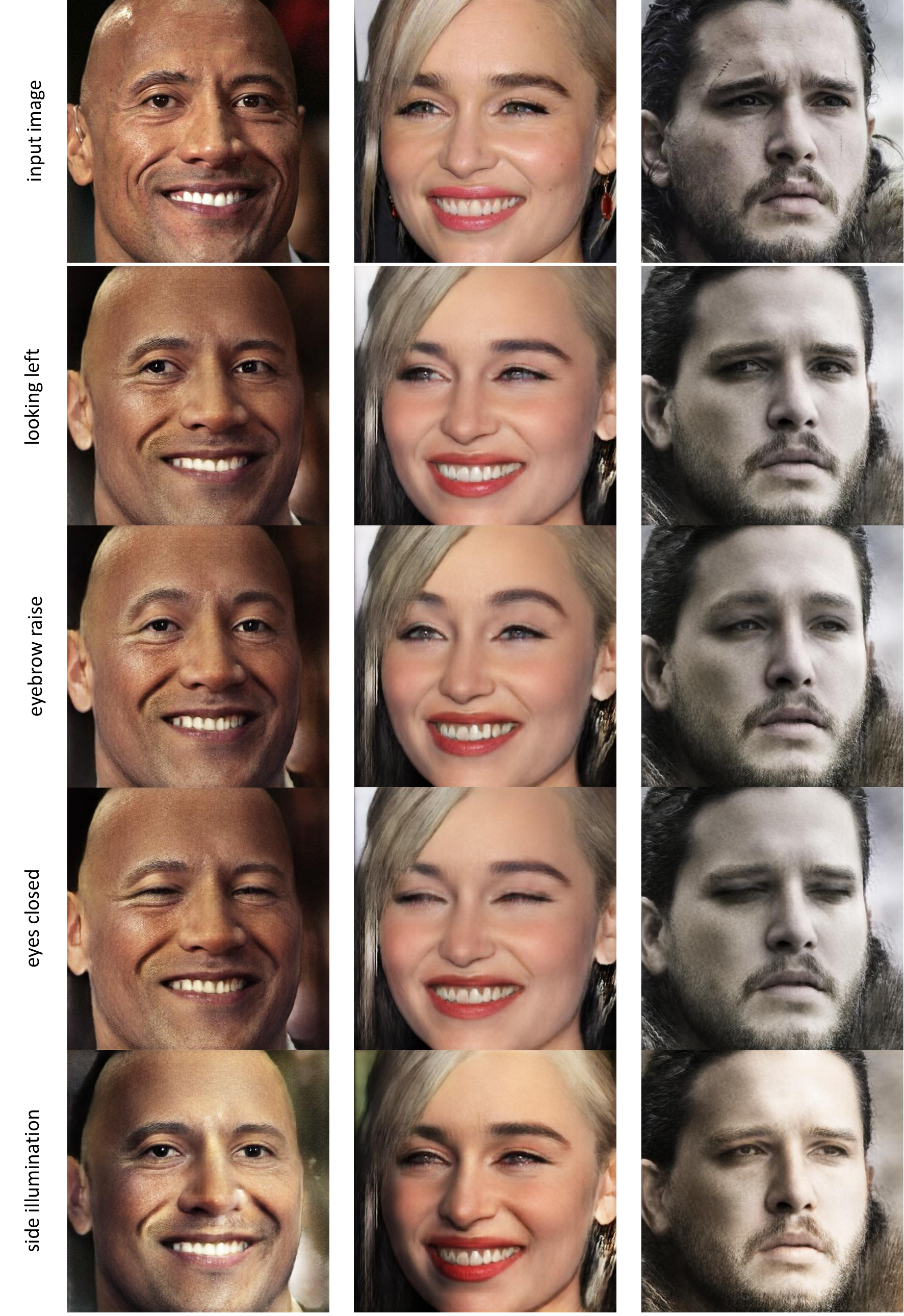}
    \caption{Results of \name{} trained at a resolution of $512 \times 512$ pixels. First row shows input image, while the remaining rows show outputs of \name{} fine-tuned on the input image. For expression modification we use the fine-grained control procedure (Section \ref{sec:fine_grained_control}) that allows to modify a subset of the expressions.}
    \label{fig:512_results}
\end{figure}
While, for simplicity and efficiency, all the experiments are performed on $256 \times 256$ pixel images, it is straightforward to increase the output resolution by adding additional layers to the generator and the image discriminators. Figure \ref{fig:512_results} shows the results of \name{} trained at a resolution of $512 \times 512$ pixels.

\subsection{Comparison to FaceID-GAN}
\begin{figure}
    \centering
    \begin{subfigure}[t]{\textwidth}
    \includegraphics[width=\textwidth]{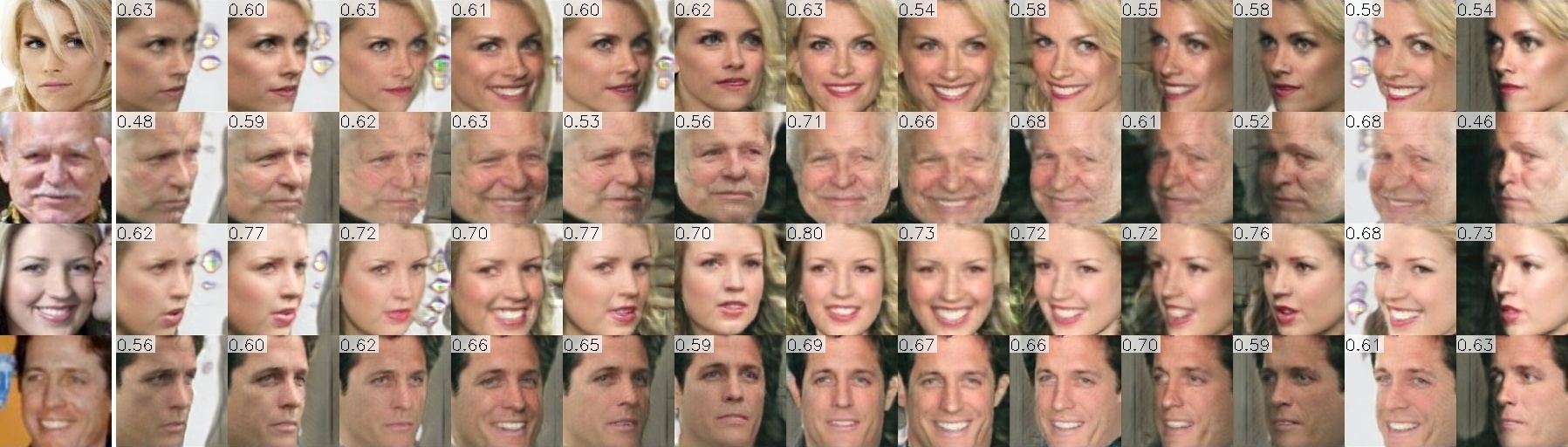}
    \caption{}
    \end{subfigure}
    \begin{subfigure}[t]{\textwidth}
    \includegraphics[width=\textwidth]{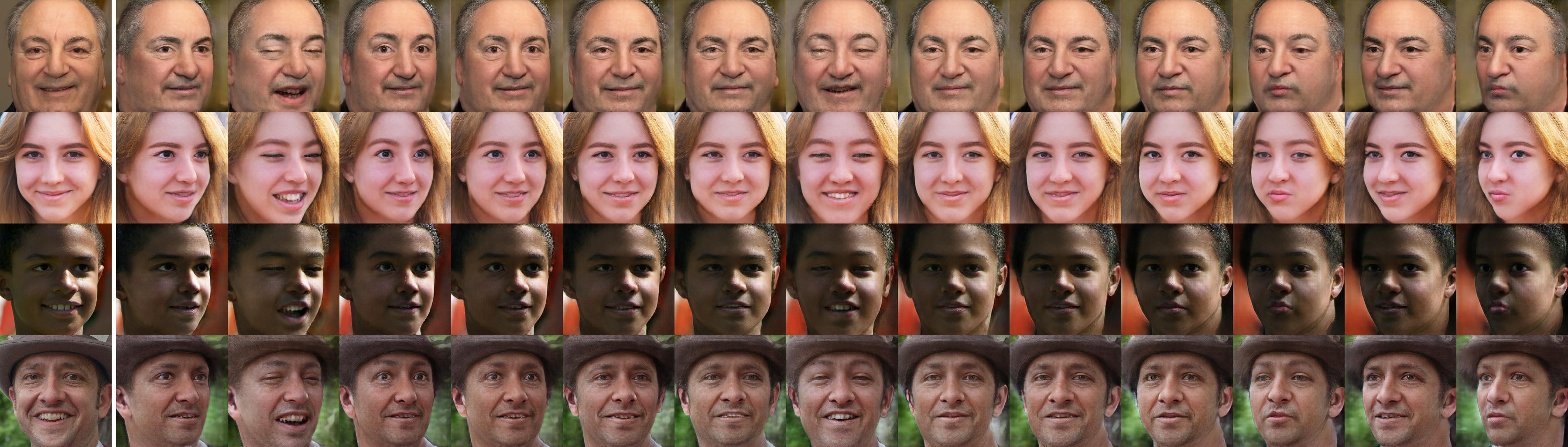}
    \caption{}
    \end{subfigure}
    \caption{Qualitative comparison of a) FaceID-GAN \cite{shen2018faceid} and b) \name{}. First column shows the input image, while the remaining columns show varying pose and expression. The identities in a) come from the CelebA \cite{CelebA} and LFW \cite{LFW} datasets, while the ones in b) come from the validation set of FFHQ \cite{StyleGAN}. Compared to FaceID-GAN, \name{} produces results with more consistent illumination and background.}
    \label{fig:faceid_gan_comparison}
\end{figure}

We perform a qualitative comparison to FaceID-GAN \cite{shen2018faceid} by fine-tuning \name{} on a number of images from the FFHQ validation set and generating a variety of poses and expressions to match a figure shown in the FaceID-GAN paper. The results of this comparison are shown in Figure \ref{fig:faceid_gan_comparison}. While both methods produce consistent poses and expression and preserve the identity well, \name{} outputs images with more consistent illumination and background. This is best exemplified in the third row of Figure \ref{fig:faceid_gan_comparison}b where the direction and intensity of illumination stay constant despite change of head pose and expression.

\subsection{Comparison to CycleGAN}
\begin{figure}
    \centering
    \includegraphics[width=0.16\textwidth]{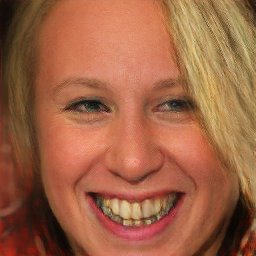}
    \hspace{-2ex}
    \includegraphics[width=0.16\textwidth]{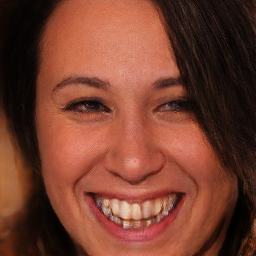}
    \includegraphics[width=0.16\textwidth]{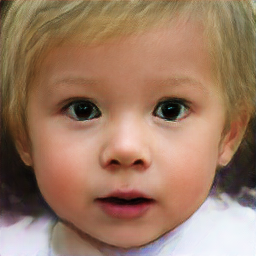}
    \hspace{-2ex}
    \includegraphics[width=0.16\textwidth]{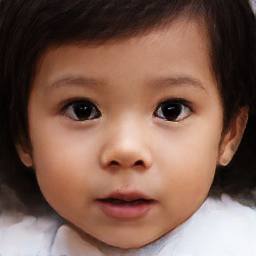}
    \includegraphics[width=0.16\textwidth]{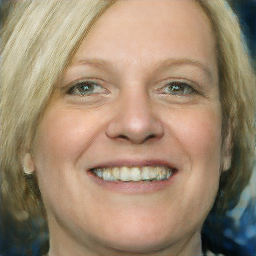}
    \hspace{-2ex}
    \includegraphics[width=0.16\textwidth]{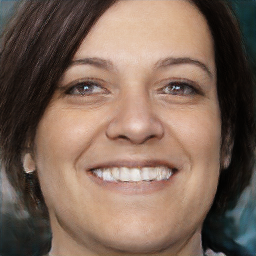}

    \includegraphics[width=0.16\textwidth]{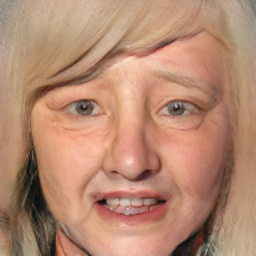}
    \hspace{-2ex}
    \includegraphics[width=0.16\textwidth]{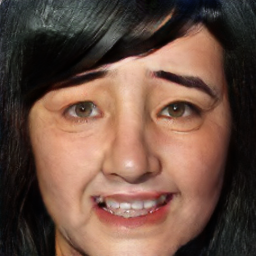}
    \includegraphics[width=0.16\textwidth]{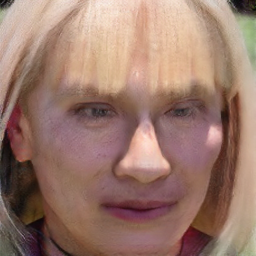}
    \hspace{-2ex}
    \includegraphics[width=0.16\textwidth]{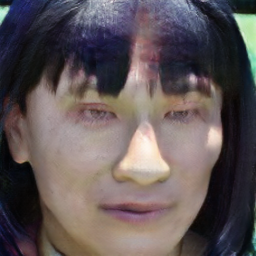}
    \includegraphics[width=0.16\textwidth]{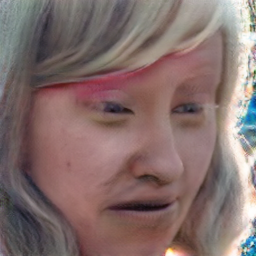}
    \hspace{-2ex}
    \includegraphics[width=0.16\textwidth]{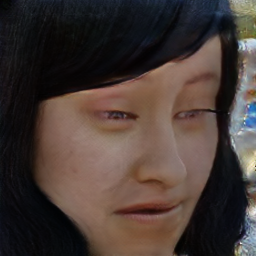}

    \caption{Change of hair colour from blond to black using \name{} (top) and synthetic data processed by CycleGAN (bottom). CycleGAN fails to preserve facial attibutes like expression and skin colour that should remain constant between the image pairs.}
    \label{fig:hair_colour_change}
\end{figure}

We train CycleGAN \cite{CycleGAN} to convert images from the synthetic domain (SynthFace \cite{SynthFace}) to the real domain (FFHQ \cite{StyleGAN}).
We then generate synthetic image pairs $I_+$ and $I_-$, where all the parameters except the modified attribute $v$ are randomly generated and identical between the two images. These images are then passed through CycleGAN and an attribute predictor trained on CelebA \cite{CelebA} to generate the controllability metrics: $C_{pred}(I_+), C_{pred}(I_-)$ and MD (Section \ref{sec:evaluation}). While CycleGAN and \name{} are not fully comparable, as the former does not allow for modifying existing real face images, this simple procedure allows us to compare the two methods in terms of the level of control over images in the real domain.

Table \ref{tab:supl_controllability} shows the controllability metric results for \name{}, CycleGAN and the synthetic data generated with the procedure mentioned above. CycleGAN obtains a slightly better dynamic range $C_{pred}(I_+) - C_{pred}(I_-)$ than \name{}, at a cost of a larger mean absolute difference of other attributes. While the difference in MD may look small in absolute terms, the relative increase compared to \name{} is 45\%. This leads us to the conclusion that while CycleGAN can preserve the very large difference between the values of the modified attribute $v$ in $I_+$ and $I_-$, it does not preserve other, more subtle, attributes that should have remained constant. Figure \ref{fig:hair_colour_change} strenghtens this conclusion by showing the effects of hair colour change performed with CycleGAN and \name{}, where the former clearly changes other face attributes as well.

The controllability metric results produced by unmodified synthetic data show lowest MD, which is expected, but also lowest dynamic range. We believe that this is caused by the domain gap between the synthetic dataset and the CelebA dataset which was used to train the attribute predictor.

To compare \name{} to CycleGAN in terms of the photorealism of generated images, we generate an additional set of 10k synthetic images and pass them through CycleGAN. We then compute the Frechet Inception Distance (FID) between the converted synthetic images and images from the FFHQ validation set. The FID score for CycleGAN is 37.74, while the score for \name{} is 33.41 as reported in Section \ref{sec:experiments}.

\begin{table}[tb]
    \caption{Average controllability metrics for \name{}, CycleGAN and synthetic data. Ideally, $C_{pred}(I_+)=1, ~C_{pred}(I_{-})=0$ and MD should be 0. The mean difference $C_{pred}(I_+) - C_{pred}(I_-)$ gives the dynamic range of a given attribute, the higher it is the more controllable the attribute.}
    \centering
    \begin{tabular}{ | l | c | c | c | c | }
      \hline
      Method & $C_{pred}(I_+)\uparrow$ & $C_{pred}(I_{-})\downarrow$ & MD$\downarrow$ & $C_{pred}(I_+) - C_{pred}(I_-)\uparrow$\\
      \hline
      base method & 0.54 & 0.04 & 0.055 & 0.50\\
      CycleGAN & 0.52 & 0.01 & 0.080 & 0.51\\
      synthetic data & 0.46 & 0.01 & 0.051 & 0.45\\
      \hline
    \end{tabular}
    \label{tab:supl_controllability}
\end{table}

\subsection{Implementation details}
For the perceptual loss we use layers conv\_1\_2, conv\_2\_2, conv\_3\_4, conv\_4\_4 of VGG-19. We regularize all the discriminators with the $R_1$ gradient penalty described in \cite{mescheder2018training}. In the image discriminators, we also use the style discriminator loss $\mathcal{L}_{style}$ described in HoloGAN \cite{HoloGAN}, while in the generator we add the identity loss $\mathcal{L}_{identity}$ described in the same paper. While HoloGAN re-uses the discriminator features for identity loss, we use a separate network that has the same architecture as the image discriminators. We do so, because neither of our discriminators is trained to work with both real and synthetic data. We set the loss weights as follows: eye loss weight $\lambda_{eye}=5$, domain adverserial loss weight $\lambda_{DA}=5$, identity loss weight $\lambda_{identity}=10$, gradient penalty loss weight $\lambda_{R_1}=10$, perceptual loss weight in 1st stage $\lambda_{perc}=0.00005$, perceptual loss weight in 2nd stage $\lambda_{perc}=0.0005$. The adveserial losses on the images and style discriminator losses all have weight $1$.

In the first training stage we sample $z \sim \mathcal{N}(0, \mathcal{I})$ and $r_R \sim \mathcal{U}(-r_{lim}, r_{lim})$, where $r_R$ is the rotation sample for real data and $r_{lim}$ is a pre-determined, per axis rotation limit. In all our experiments we set $r_{lim}$ to be identical to the rotation limits used in synthetic data generation as described in the dataset section. In the second stage the $E_R$ output corresponding to $r_R$ is constrained to the range specified in $r_{lim}$ by using a tanh activation and multiplying the output by $r_{lim}$.

Table \ref{tab:generator_arch} shows the architecture of the generator network $G$. In each AdaIN \cite{huang2017arbitrary} input the latent vector $z$ is processed by a 2-layer MLP. The volume rotation layer is the same as the one used in HoloGAN \cite{HoloGAN}. Table \ref{tab:discriminator_arch} shows the architecture of the image discriminators $D_R$, $D_S$. Following \cite{HoloGAN}, most of the convolutional layers of the discriminator use instance normalization \cite{ulyanov2016instance}. The latent GAN generator $G_{lat}$ and discriminator share the same 3-layer MLP architecture.

The networks are optimized using Adam \cite{Adam} with a learning rate of 4e-4. We perform the first stage of training for 50k iterations and then the second stage for 100k iterations. The latent GAN is also trained for 100k iterations. Following \cite{ProgressiveGAN}, in both the latent GAN and decoder $G$, we keep an exponential running mean of the weights during training and use those smoothed weights to generate all results.

\begin{table}[t!]
    \caption{Architecture of the generator network $G$}
    \centering
    \begin{tabular}{ l c c c c}
        Layer name & Kernel shape & Activation & Output shape & Normalisation \\
        \hline
        learned const input & - & - & $4 \times 4 \times 4 \times 512$ & - \\
        upsampling & - & - & $8 \times 8 \times 8 \times 512$ & - \\
        conv3d\_1 & $3 \times 3 \times 3$ & LReLU & $8 \times 8 \times 8 \times 256$ & AdaIN \\
        upsampling & - & - & $16 \times 16 \times 16 \times 256$ & - \\
        conv3d\_2 & $3 \times 3 \times 3$ & LReLU & $16 \times 16 \times 16 \times 128$ & AdaIN \\
        volume rotation & - & - & $16 \times 16 \times 16 \times 128$ & - \\
        conv3d\_3 & $3 \times 3 \times 3$ & LReLU & $16 \times 16 \times 16 \times 64$ & - \\
        conv3d\_4 & $3 \times 3 \times 3$ & LReLU & $16 \times 16 \times 16 \times 64$ & - \\
        reshape & - & - & $16 \times 16 \times (16 \cdot 64)$ & - \\

        conv2d\_1 & $1 \times 1$ & LReLU & $16 \times 16 \times 512$ & - \\
        conv2d\_2 & $4 \times 4$ & LReLU & $16 \times 16 \times 256$ & AdaIN \\
        upsampling & - & - & $32 \times 32 \times 256$ & - \\

        conv2d\_3 & $4 \times 4$ & LReLU & $32 \times 32 \times 64$ & AdaIN \\
        upsampling & - & - & $64 \times 64 \times 64$ & - \\

        conv2d\_4 & $4 \times 4$ & LReLU & $64 \times 64 \times 32$ & AdaIN \\
        upsampling & - & - & $128 \times 128 \times 32$ & - \\

        conv2d\_5 & $4 \times 4$ & LReLU & $128 \times 128 \times 32$ & AdaIN \\
        upsampling & - & - & $256 \times 256 \times 32$ & - \\
        conv2d\_6 & $4 \times 4$ & tanh & $256 \times 256 \times 3$ & - \\

    \end{tabular}
    \label{tab:generator_arch}
\end{table}
\addtolength{\tabcolsep}{2pt}
\begin{table}[t!]
    \caption{Architecture of the image discriminator networks $D_R$, $D_S$}
    \centering
    \begin{tabular}{ l c c c c}
        Layer name & Kernel shape, stride & Activation & Output shape & Normalisation \\
        \hline
        conv2d\_1 & $1 \times 1, 1$ & - & $256 \times 256 \times 3$ & - \\
        conv2d\_2 & $3 \times 3, 2$ & LReLU & $128 \times 128 \times 48$ & Instance Norm \\
        conv2d\_3 & $3 \times 3, 2$ & LReLU & $64 \times 64 \times 96$ & Instance Norm \\
        conv2d\_4 & $3 \times 3, 2$ & LReLU & $32 \times 32 \times 192$ & Instance Norm \\
        conv2d\_5 & $3 \times 3, 2$ & LReLU & $16 \times 16 \times 384$ & Instance Norm \\
        conv2d\_6 & $3 \times 3, 2$ & LReLU & $8 \times 8 \times 768$ & Instance Norm \\
        fully connected & $49152$ & - & $1$ & - \\
    \end{tabular}
    \label{tab:discriminator_arch}
\end{table}
\addtolength{\tabcolsep}{-2pt}

\subsection{Factorized latent space details}
\begin{table}[t!]
    \caption{Dimensionalities and descriptions of latent space factors}
    \centering
    \begin{tabular}{ l c c c c}
        Factor name & $\dim{\theta_i}$ & $\dim{z_i}$ & Description of $\theta_i$\\
        \hline
        beard style & 9 & 7 & PCA coefficients \\
        eyebrow style & 44 & 7 & PCA coefficients \\
        expression & 52 & 30 & 3D head model parameters $\in [0, 1]$ \\
        eye colour & 6 & 3 & one-hot encoding \\
        eye rotation & 3 & 2 & rotation angles \\
        hair colour & 3 & 3 & melanin, grayness, redness\\
        hair style & 18 & 8 & PCA coefficients \\
        head shape & 53 & 30 & 3D head model parameters\\
        illumination & 50 & 20 & PCA coefficients\\
        lower eyelash style & 3 & 2 & one-hot encoding\\
        texture & 50 & 30 & VAE latent space vector\\
        upper eyelash style & 3 & 2 & one-hot encoding\\
    \end{tabular}
    \label{tab:factorization}
\end{table}

Table \ref{tab:factorization} shows the dimensionalities of all latent space factors $z_i$ and corresponding synthetic data parameters $\theta_i$. The dimensionalities of each $z_i$ were chosen based on perceived complexity of the feature, for example we allocate more dimensions to expression than to hair colour. We are able to exert control over all the parameters with the exception of eyelash styles, which correspond to features that are too small at the image resolution we are working in. The expression parameters consist of the 51 expression blendshapes described in \cite{SynthFace} and one additional dimension for the rotation of the jaw bone that leads to mouth opening.

\subsection{Controllability metric details}
The attribute predictor $C_{pred}$ we use for the metrics is a MobilenetV2 \cite{sandler2018mobilenetv2} trained to predict 38 of CelebA's 40 attributes. The two attributes we do not predict are Wearing\_Necklace and Wearing\_Necktie as the required features are not present in our crops of CelebA images. For controllability metrics computed using the attribute predictor we use \name{} to drive 8 attributes, while we use all 38 attributes to compute the MD value. The 8 attributes we drive are chosen to be non-ambiguous and easy to verify by the user study participants.

For each evaluated face attribute $v$ we set the corresponding $z_i=E_{S_i}(\theta_i)$, where $\theta_i$ is determined by manual inspection for both $I_-$ and $I_+$. For example, for $v$ smile we set the expression parameters that correspond to mouth corners going up for $I_+$, while for $I_-$ we set parameters that correspond to mouth corners going down. Note that in this experiment we do not use the fine-grained control method described in Section \ref{sec:fine_grained_control}, we instead set the entire $\theta_i$ with the chosen value.

\subsection{Entanglement in synthetic dataset}
\begin{figure}
    \centering
    \includegraphics[width=0.97\textwidth]{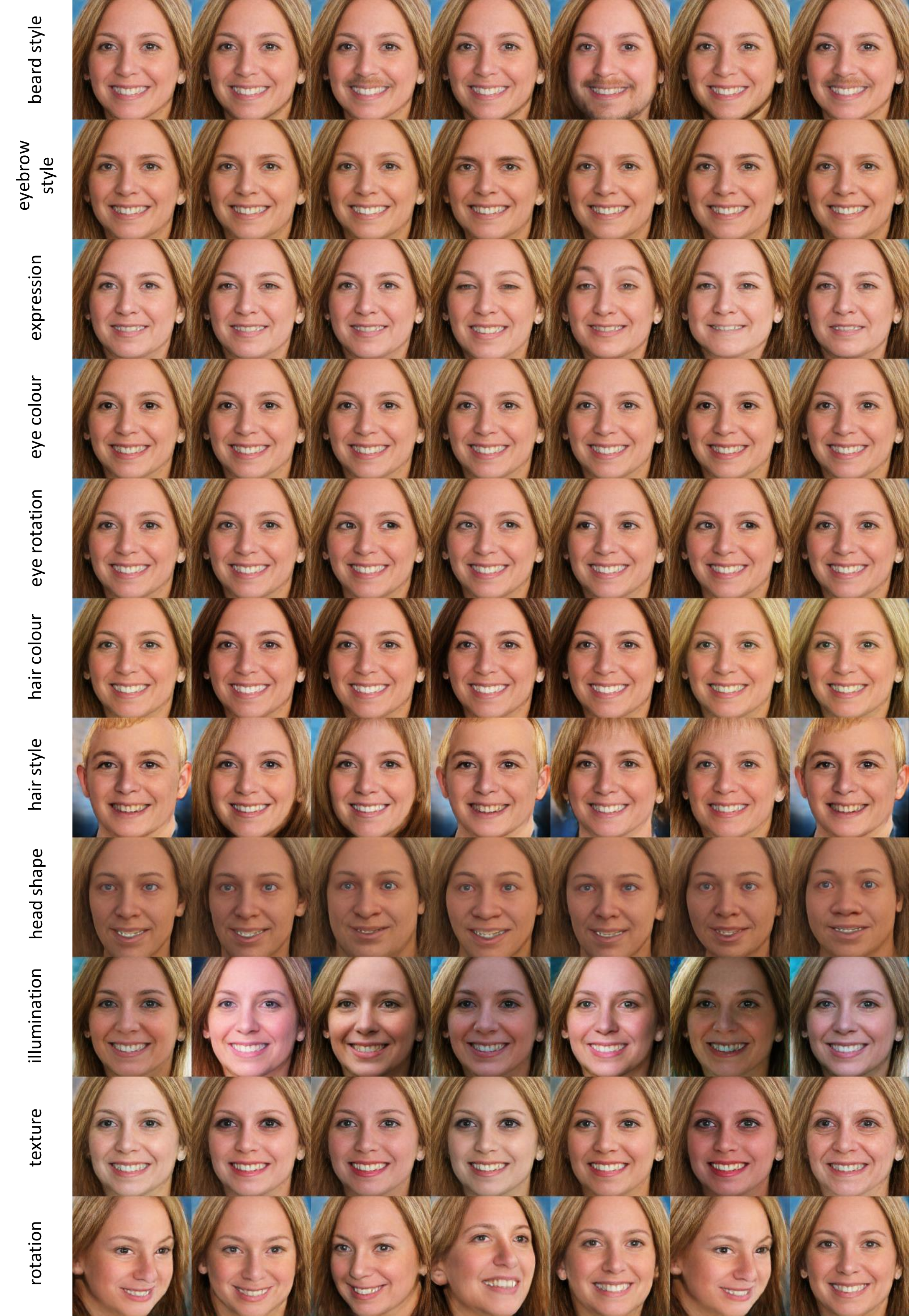}
    \caption{Random variations of the input factors of \name{} shown on a single base image. In each row a single latent space factor $z_i=E_S(\theta_i)$ is modified, with each row showing a different, randomly selected $\theta_i$. Note the change of appearance when modifying head shape, discussion in Section \ref{sec:failure_modes}.}
    \label{fig:all_attribute_variation}
\end{figure}

While \synthdatasetname{} allows for disentangling many face attributes, we have found that some of it's properties lead to entanglement. In \synthdatasetname{} each texture is applied together with a corresponding displacement map, this leads to entanglement between texture and face shape. Examples of such entanglement are noticeable in texture row of Figure \ref{fig:all_attribute_variation}. The eyebrow style row of the same figure shows changes in eyebrow height, which are entangled with the eyebrow raise expression. This is due to the varying vertical placement of eyebrows in \synthdatasetname{}.

Another issue is that in \synthdatasetname{} a neutral face can have an open mouth. Because of that, applying a neutral expression does not always lead to the mouth closing, this issue is visible in top row of Figure \ref{fig:puppetgan_comp_mouth_opening}. In those cases, the mouths can still be closed by setting the value of the mouth opening expression to negative. The same issue applies to other expressions to a smaller degree.

\subsection{Additional figures}
\begin{figure}
    \centering
    \includegraphics[width=\textwidth]{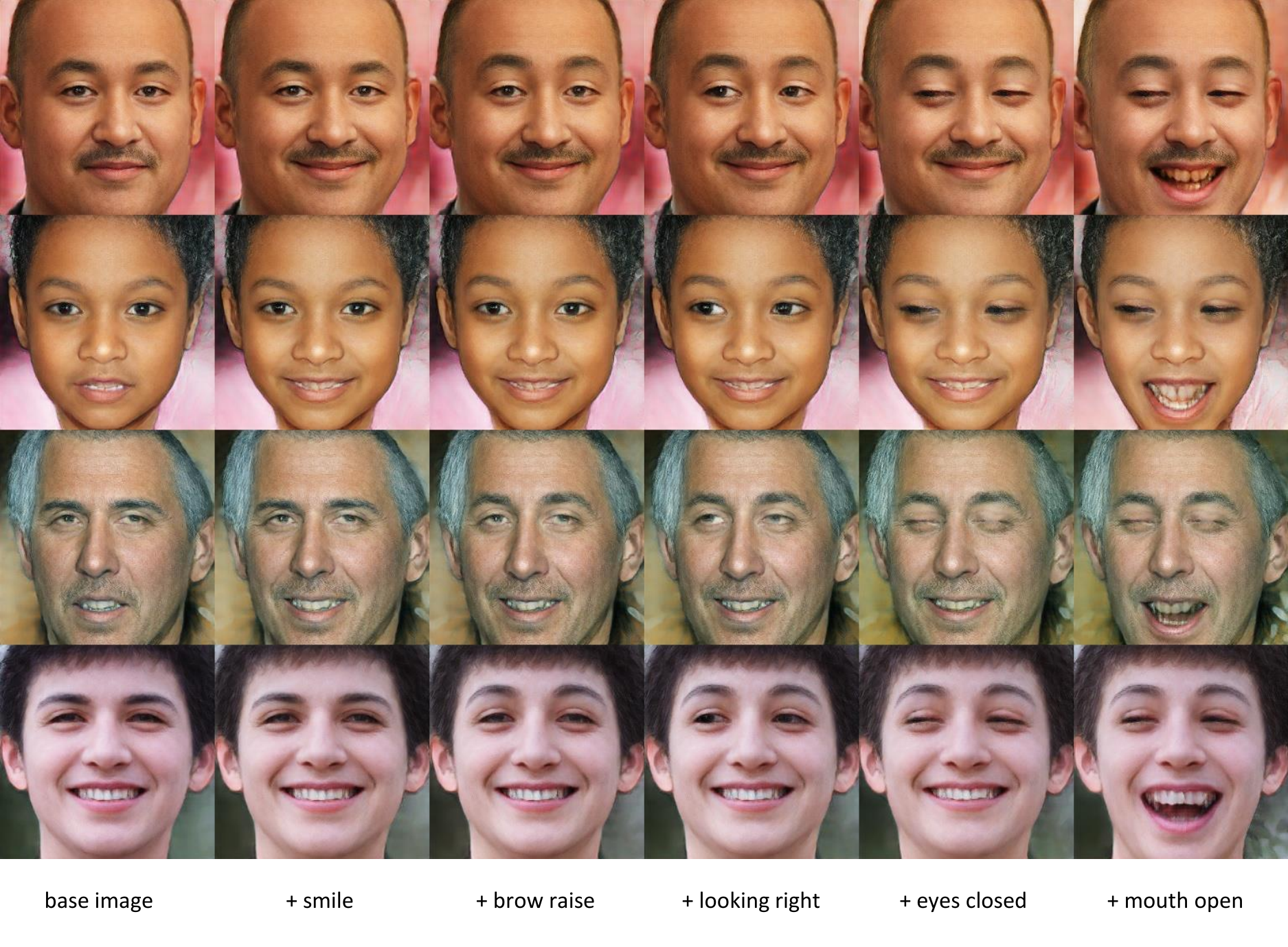}
    \caption{Generation of combinations of facial expressions. The left column shows the base image while each subsequent column adds another expression to the image. With the exception of looking right each expression is added using the fine-grained control method described in Section \ref{sec:fine_grained_control}.}
    \label{fig:fine_grained_combinations}
\end{figure}

\begin{figure}
    \centering
    \includegraphics[width=\textwidth]{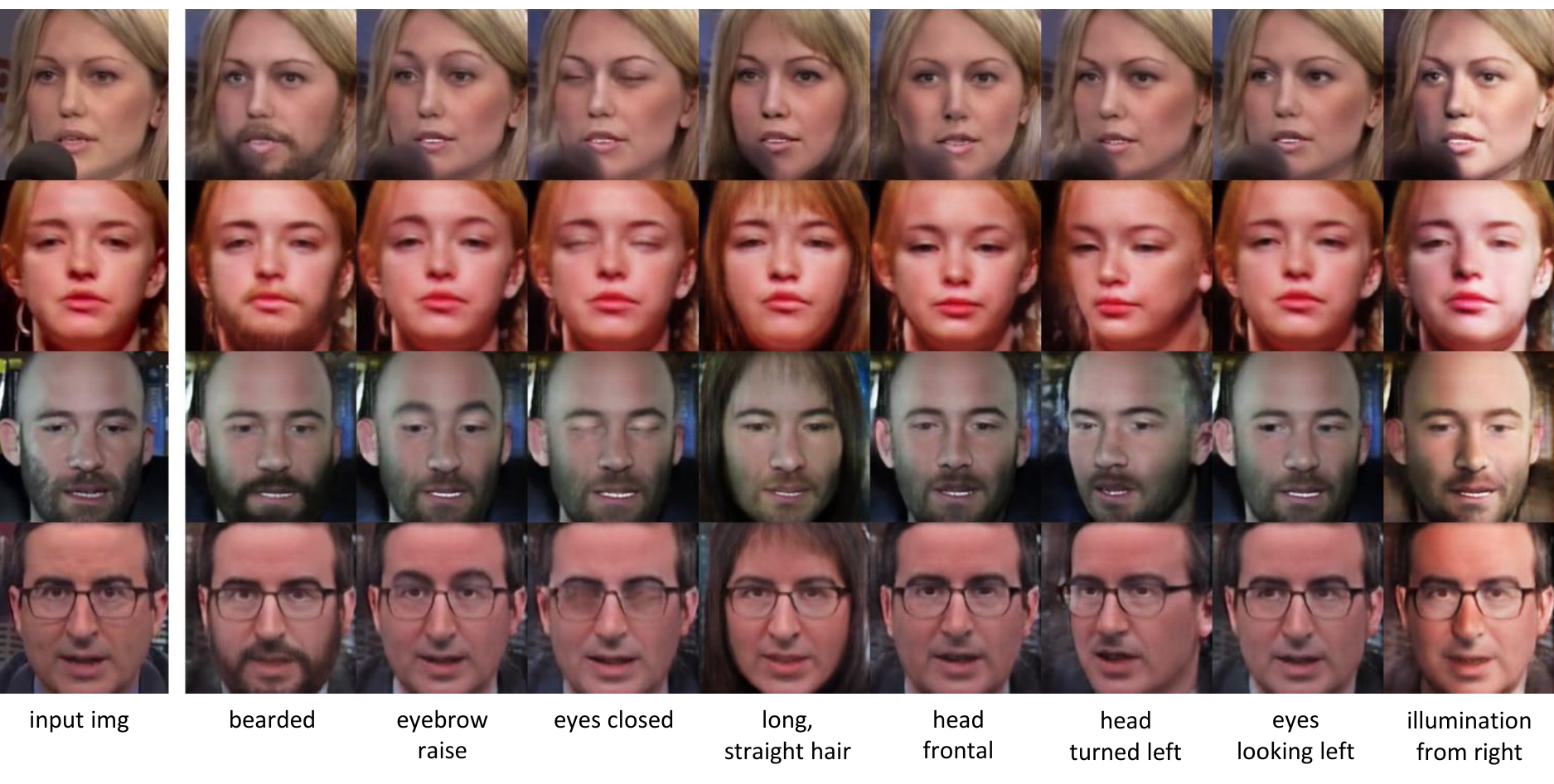}
    \caption{Modification of various attributes of faces from the 300-VW dataset \cite{300VW} using \name{} fine-tuned on the input image.}
    \label{fig:puppetgan_comp_other_variations}
\end{figure}

\begin{figure}
    \centering
    \includegraphics[width=\textwidth]{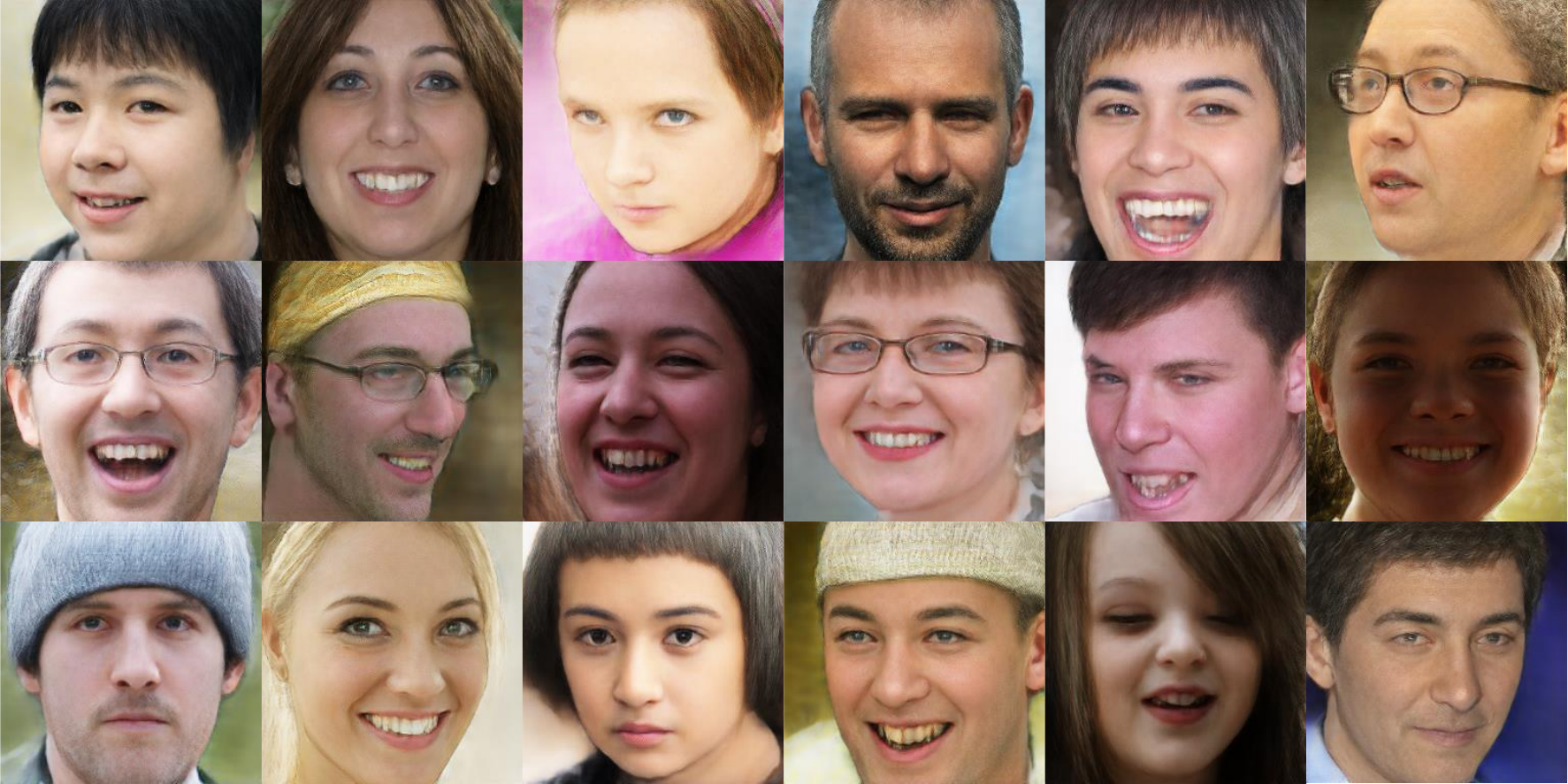}
    \caption{Results of sampling the latent space $z=E_R(I_R)$, where $I_R$ is randomly selected from FFHQ \cite{StyleGAN}.}
\end{figure}

\begin{figure}
    \centering
    \begin{subfigure}[t]{\textwidth}
        \includegraphics[width=\textwidth]{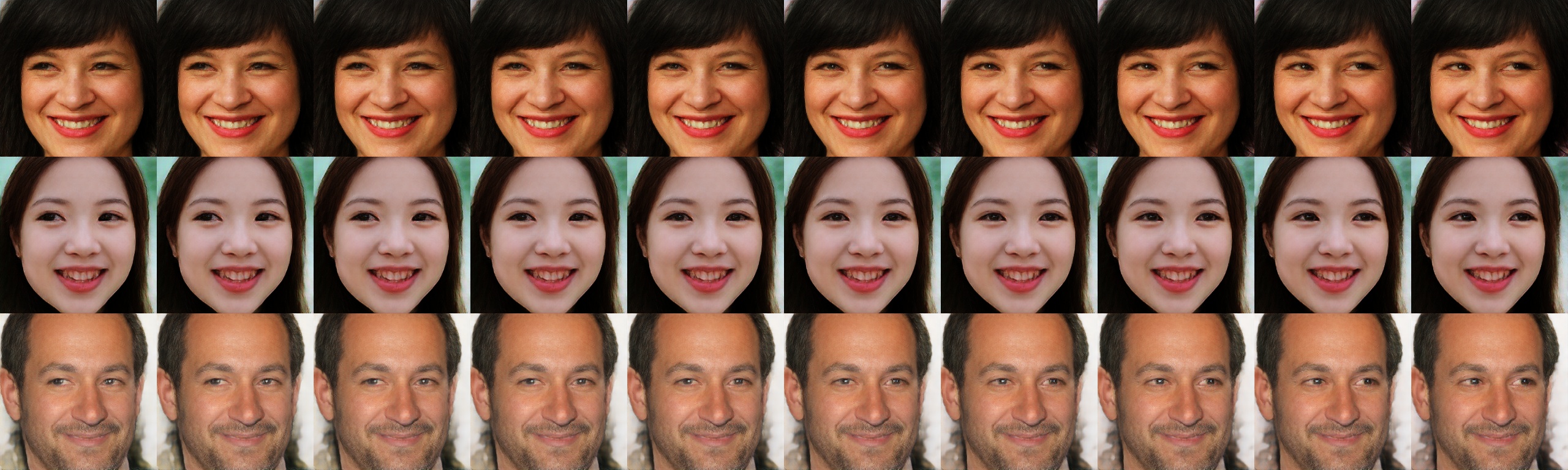}
        \caption{Model trained with the eye gaze loss}
    \end{subfigure}
    \begin{subfigure}[t]{\textwidth}
        \includegraphics[width=\textwidth]{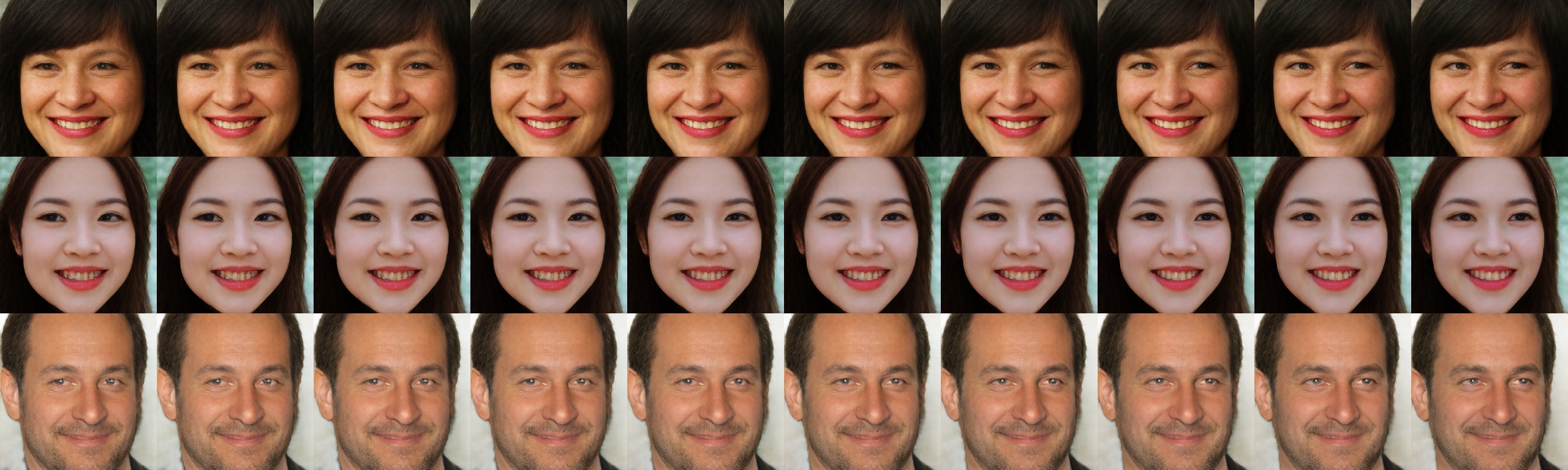}
        \caption{Model trained without the eye gaze loss}
    \end{subfigure}
    \caption{Comparison of control over eye gaze direction for a model trained with the eye gaze loss $\mathcal{L}_{eye}$ and a model trained without it. Notice that a) achieves a wider range of eye motion, which is most visible in the first and last column of each row.}
    \label{fig:eye_gaze_loss}
\end{figure}

\begin{figure}
    \centering
    \includegraphics[width=0.45\textwidth]{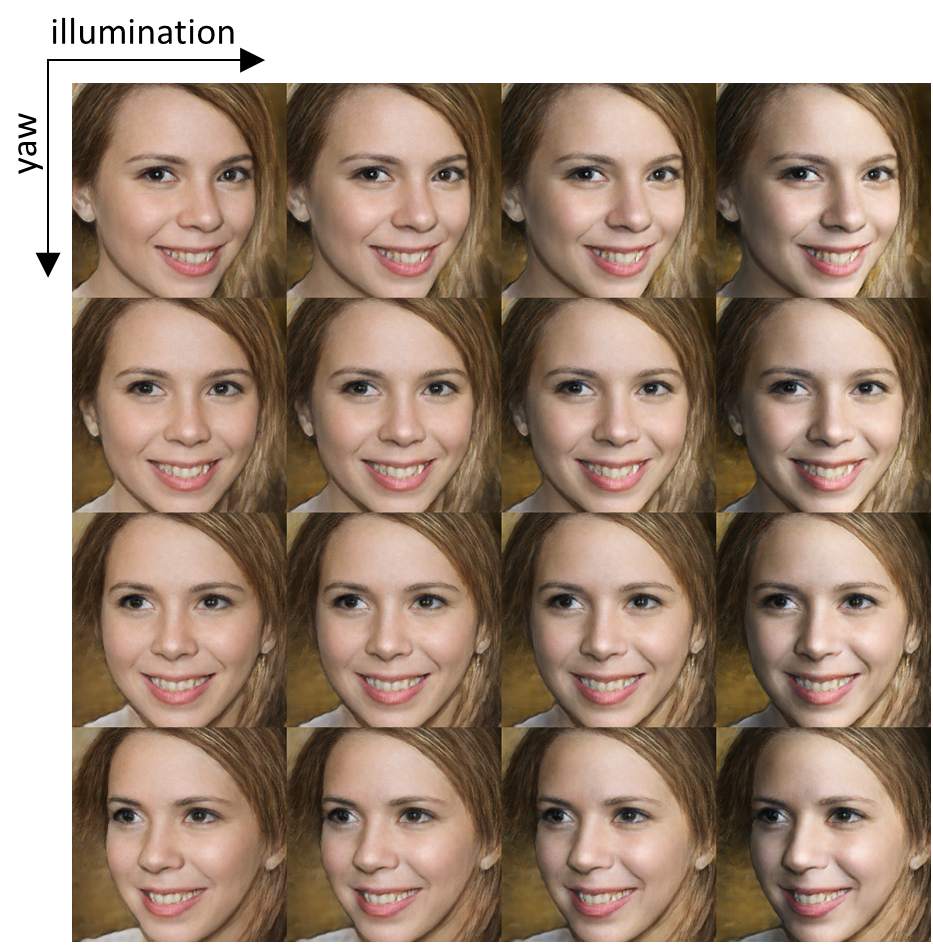}
    \quad
    \includegraphics[width=0.45\textwidth]{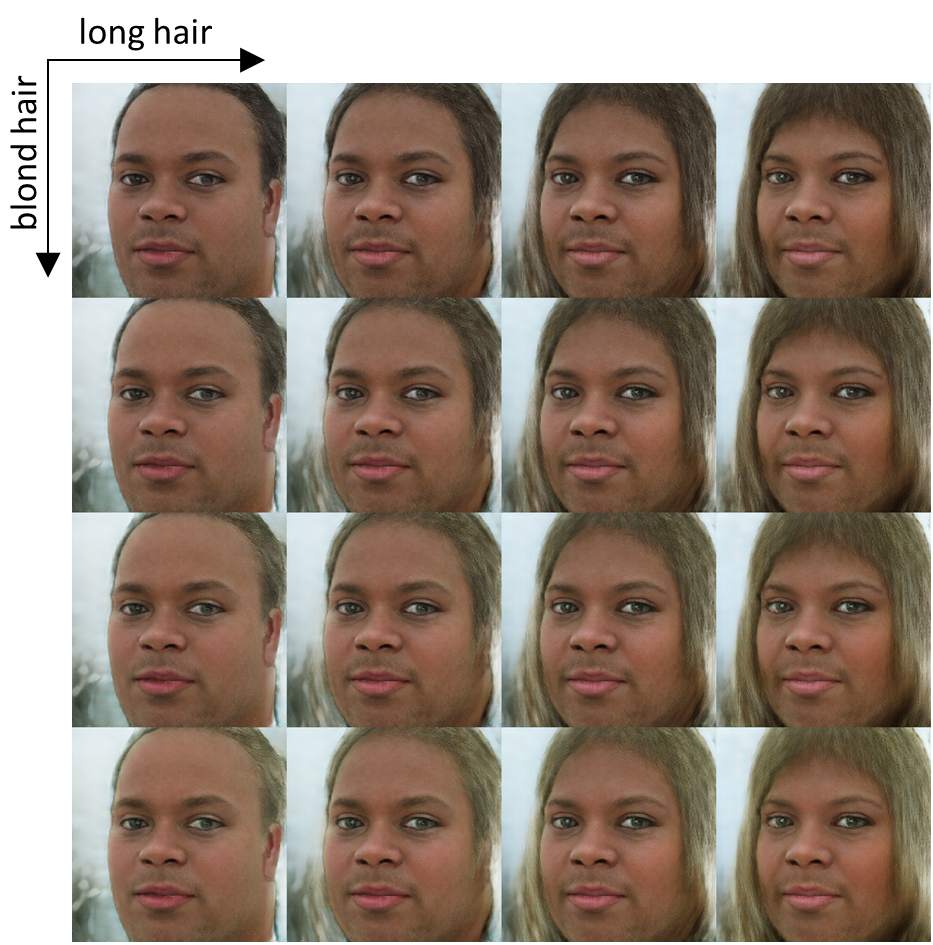}
    \caption{Interpolation of two face attribute at a time in images sampled from the FFHQ \cite{StyleGAN} dataset. Note that the illumination direction stays constant as the head rotates.}
\end{figure}

\begin{figure}
    \centering
    \includegraphics[width=\textwidth]{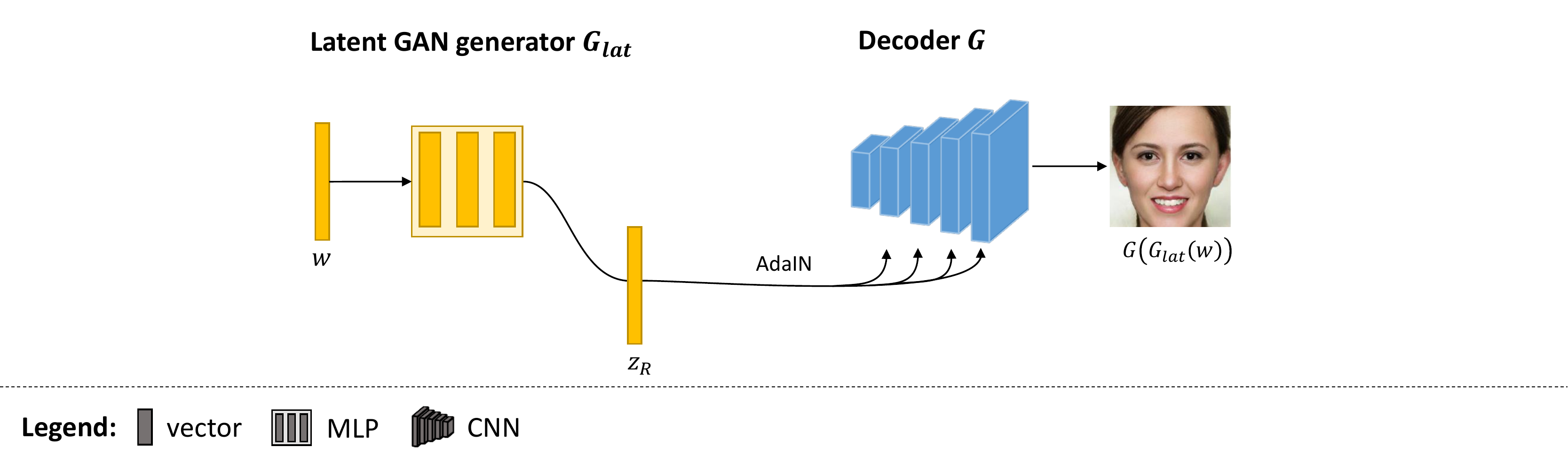}
    \caption{Outline of the method at inference time, when the latent GAN (described in Section \ref{sec:latent_gan}) is used to sample the latent space $z_r=G_{lat}(w)$, where $w \sim \mathcal{N}(0, \boldsymbol{I})$. Keep in mind that $z_r$ can still be modified using any of the methods described in the paper to achieve control over the output image.}
    \label{fig:outline_with_Glat}
\end{figure}

\begin{algorithm}[H]
    \SetAlgoLined
    \KwIn{latent space factor $z_i$, number of iterations $n$, learning rate $\gamma$}
    \KwResult{modified latent space factor $z_i$}
     iter = 0\;
     $\tilde{\theta_i}=[0, 0, \dots, 0]$\;
     \While{$iter<n$}{
      $L = \vert z_i - E_{S_i}(\tilde{\theta}_i)\vert^2$\;
      $\tilde{\theta_i} = \tilde{\theta_i} - \gamma \triangledown L$\;
      clip $\tilde{\theta_i}$ to the range of valid values of $\theta_i$\;
      iter++\;
    }
    apply desired modification to $\tilde{\theta_i}$\;
    $z_i=E_{S_i}(\tilde{\theta}_i)$\;

     \caption{Fine grained control. The algorithm uses projected gradient descent to estimate the synthetic data rendering parameter $\tilde{\theta_i}$ that maps to the supplied $z_i$. $\tilde{\theta_i}$ can then be modified to obtain a new $z_i=E_S(\tilde{\theta_i})$. Since $\tilde{\theta_i}$ is a parameter of a computer graphics pipeline, it is semantically meaningful and its individual components can be modified to achieve fine-grained control.}
     \label{alg:fine_grained}
\end{algorithm}